\def\year{2018}\relax
\documentclass[letterpaper]{article} 
\usepackage{aaai18}  
\usepackage{times}  
\usepackage{helvet}  
\usepackage{courier}  
\usepackage{url}  
\usepackage{graphicx}  
\usepackage{booktabs} 
\usepackage{bm}
\usepackage[ruled,linesnumbered]{algorithm2e}
\usepackage{subfigure}
\usepackage{multirow}
\usepackage{algpseudocode}
\usepackage{amsmath,amssymb,amsthm}
\begin{document}
%
\title{Group Sparse Bayesian Learning for Active Surveillance 
on Epidemic Dynamics}

\author{Hongbin Pei$^{1,2}$ \and Bo Yang$^{1,2}$ \and Jiming Liu$^{3}$ \and Lei Dong$^{4}$ \\
$^{1}$College of Computer Science and Technology, Jilin University, Changchun, China \\
$^{2}$Key Laboratory of Symbolic Computation and Knowledge Engineering of Ministry of Education, China\\
$^{3}$Department of Computer Science, Hong Kong Baptist University, Hong Kong \\
$^{4}$Institute of Remote Sensing and Geographical Information Systems, Peking University, Beijing, China\\
peihb15@mails.jlu.edu.cn}

\maketitle
\begin{abstract}
Predicting epidemic dynamics is of great value in understanding and controlling diffusion processes, such as infectious disease spread and information propagation. 
This task is intractable, especially when surveillance resources are very limited. 
To address the challenge, we study the problem of active surveillance, i.e., how to identify a small portion of system components as sentinels to effect monitoring, such that the epidemic dynamics of an entire system can be readily predicted from the partial data collected by such sentinels. 
We propose a novel measure, the $\bm{\gamma}$ value, to identify the sentinels by modeling a sentinel network with row sparsity structure.
We design a flexible group sparse Bayesian learning algorithm to mine the sentinel network suitable for handling both linear and non-linear dynamical systems by using the expectation maximization method and variational approximation. 
The efficacy of the proposed algorithm is theoretically analyzed and empirically validated using both synthetic and real-world data.
\end{abstract}

\section{1 ~ Introduction}
Predicting epidemic dynamics is of great value in understanding and controlling diffusion processes.
Diffusion phenomena, such as infectious disease spread and information propagation, exist widely in the real world.
By the notion of a dynamical system, which is a powerful tool for characterizing dynamics \cite{brunton2016discovering}, the task of epidemic dynamics prediction is to estimate ensuing system states using given current states. 
Therefore, the foundation of this task is surveillance, which is to monitor and report the current system states in a timely manner.

However, in practice, it is often very challenging to monitor the overall components in a system because diffusion phenomena often cross over very large spatiotemporal ranges, e.g., spreading of infectious diseases  in a country \cite{polgreen2009optimizing}, air contaminant diffusion in a large city \cite{zheng2013uair}, and hot topics/meme forwarding on social media \cite{chen2013emerging}.
For large-scale diffusion phenomena, timely and all-around monitoring is hard and infeasible, especially when available surveillance resources are very limited. 

Owing to a lack of systematic deployment of already limited surveillance resources, disease surveillance has long suffered from low reporting rates, biased sampling, and lengthy reporting time-lags \cite{gerardo2013structuring}.
For instance, Tengchong city, a malaria endemic region in China, has 18 towns (consisting of 221 villages), 167,964 households, and 658,207 residents that are distributed in a mountainous area of 5,845 square kilometers.  From 2005 to 2011 in Tengchong, 7,835 confirmed malaria cases were reported, but the
Tengchong Centers for Disease Control (CDC), the local disease surveillance team, could only afford to have a few staff members conducting the time-consuming case surveys.

Active surveillance is a promising strategy to address the challenge of limited surveillance resources, in which epidemic dynamics is predicted by proactively monitoring a relatively small number of sentinel components, whose state data are collected to achieve a good trade-off between prediction accuracy and surveillance cost.
The key to implementing active surveillance is to determine, in a dynamical system, which components are important for dynamics prediction and how to identify them.
This is a non-trivial task because the interaction structure among components, which characterizes the dynamics mechanism, is usually hidden and heterogenous, such as the social contact network in charge of disease spread \cite{Bo2016Characterizing}. 

Here, we address this challenge by proposing a novel importance measure, the $\bm{\gamma}$ value, to determine how important a component is to predicting epidemic dynamics.
Based on the measure, we develop a backward-selection algorithm designated the sentinel network mining algorithm (SNMA for short) to mine a sentinel network.
The sentinel network encodes the influence relationship from sentinel components to the overall system components, which is a row sparse network that contains only the influential links emitted from the sentinels.
With the discovered sentinel network, one can predict the future overall system dynamics by only monitoring and feeding the sentinels' states into a system function.

Different from existing works, we model the task of sentinel network mining as a group sparse learning problem and propose an effective and flexible Bayesian learning algorithm for various dynamical systems, including the most widely used linear continuous system and logistic discrete system.
The expectation-maximization (EM) and variational approximation methods are employed to infer the posterior distribution of the sentinel network.
Particularly, we solved the scalability problem by means of proposing more efficient multiplication and inverse operations on diagonal-block metrics.
Validations and comparisons were performed on both synthetic and real-world data, which show that the proposed method outperforms existing methods.

We summarize the main contributions of this paper as follows: (1) We propose a novel measure, the $\gamma$ value, to identify sentinel components by modeling the sentinel network with row sparsity structure;
(2) We design an effective and flexible group sparse Bayesian learning algorithm to discover the sentinel network;
(3) We solve the scalability and generality problem of this algorithm to a certain degree;
(4)We develop a comprehensive framework for active surveillance and validate it by performing extensive experiments on both synthetic and real-world data.

\section{2 ~ Related works}
In the literature, most of the optimal sensor placement studies identify the sentinels' location via the framework of the budgeted maximum coverage problem (BMCP) \cite{khuller1999budgeted}, which aims to maximize a specific objective by finding a set of components with the minimum budget (the budget is often defined as the set size). 
Examples include the timely detection of contaminated water \cite{krause2008efficient} and early detection of the outbreak of Weblogs \cite{leskovec2007cost}.
As BMCP is NP-hard, heuristic algorithms like sub-modular maximization are often employed. 
Note that designing the objective functions of BMCP relies heavily on the known interaction structure among components. 
When the underlying interaction structure cannot be observed directly, such as the social media network in charge of information diffusion and the social contact network for disease spread, the current methods proposed within the BMCP framework cannot be used  \cite{gomez2010inferring}.

In spatial statistics, Gaussian processes (GPs) can effective represent the spatial correlation and uncertainty of the sensed field.
Based on GPs, one can adopt general information criteria, typically such as mutual information  (GPs-MI), to greedily select sentinels \cite{krause2008near,hoang2014nonmyopic}. 
In this framework, both the GPs and the information criteria are model-free. They neglect the mechanism of data generation.
As a result, the prior knowledge of the phenomena [e.g., the susceptible-infectious-recovered (SIR) model for infectious disease \cite{Mathematical2010mathematical}] is difficult to integrate into the method.
If such available prior knowledge can be adequately incorporated, the performance of learning and prediction will be significantly improved, as shown in our experiments.

In addition to the epidemic dynamics data, some works turn to other types of available data to help identify the sentinels.
For instance, socio-economic data is leveraged to estimate the malaria infection risk caused by imported cases, and further to implement an effective active surveillance plan \cite{yang2014aaai}.
Traffic data and environmental data are used to infer real-time air quality, and further determine the best deployment locations of new air contaminant monitoring stations \cite{hsieh2015inferring}.
It is difficult to reuse these customized methods on other types of active surveillance applications, if the domain data or domain knowledge they require are not available.

\section{3 ~ Active surveillance framework}
For epidemic dynamics, we now propose the framework of active surveillance. It consists of three main steps.

\textbf{Step 1:} collect epidemic dynamics data in $N$ components of interest.

\textbf{Step 2:} mine the sentinel network from the data. In the network, the number of sentinel components (i.e., sentinel nodes) $k$ is according to a budget.

\textbf{Step 3:} with the  sentinel network, predict future epidemic dynamics of the $N$ components based on the data collected from the $k$ sentinel components.

The last two steps constitute the foundation of the framework, and we will elaborate them in following sections.

\subsection{3.1 ~ Problem formulation}
Consider a diffusion among $N$ components in a dynamical system.
Let matrix $\mathbf{D}$ $\in$ $\mathbb{R}^{T \times N}$=$[\mathbf{D}_{1},\cdots,\mathbf{D}_{T}]^{\rm T}$ be the epidemic dynamics during a time window $[1,T]$.
Specifically, $\mathbf{D}_t$=$[\mathbf{D}_{t,1},\cdots,\mathbf{D}_{t,N}]$, where each entry $\mathbf{D}_{t,i}$ denotes the state of component $i$ at time $t$, and it may be a real number (e.g., the number of newly infected cases in the city $i$) or a Boolean value (e.g., whether a news is posted in the blog $i$).
Let $\mathbf{D}^{\bm{s}}$ $\in$ $\mathbb{R}^{T \times N}$denote the surveillance data collected by $k$ sentinel components. 
Specifically, $\mathbf{D}^{\bm{s}}_{t,i}$ is equal to $\mathbf{D}_{t,i}$ when component $i$ is a sentinel, and empty otherwise. 

Let $ f(\mathbf{D}^{\bm{s}}_{t}; \mathbf{S})$ be the dynamical system function achieving the dynamics prediction.
Let matrix $\mathbf{S}$ $\in$ $\mathbb{R}^{N \times N}$ denote a sentinel network, a set of key parameters in the system function.
It depicts the influential relationship from the sentinels to all components, where each link $\mathbf{S}_{i,j}$ encodes the effect of sentinel $i$ on component $j$ by its weight.
Thus, $\mathbf{S}$  is a row sparse matrix only containing links emitted from the $k$ sentinels.
Now, the active surveillance can be formulated as to predict the future components' states based on the surveillance data $\mathbf{D}^{\bm{s}}$ and the sentinel network $\mathbf{S}$:
\begin{equation}
\footnotesize 
\mathbf{D}_{t+1} \approx \hat{\mathbf{D}}_{t+1}  = f (\mathbf{D}^{\bm{s}}_{t}; \mathbf{S}).
\label{equ:prediction_pro}
\end{equation}

From Eq. \ref{equ:prediction_pro}, two computational issues need to be addressed for the goal of active surveillance:

I) Sentinel identification: How to identify the sentinels from all components and mine the sentinel network $\mathbf{S}$ according to a given budget from the dynamics $\mathbf{D}$?

II) Sentinel prediction: How to predict the future dynamics $\mathbf{D}_{t+1}$ from the current surveillance data $\mathbf{D}^{\bm{s}}_{t}$ based on the discovered $\mathbf{S}$?
 
\subsection{3.2 ~ Sentinel identification}
\begin{figure}[!htb]
\centering{}
\includegraphics[width=0.31 \textwidth]{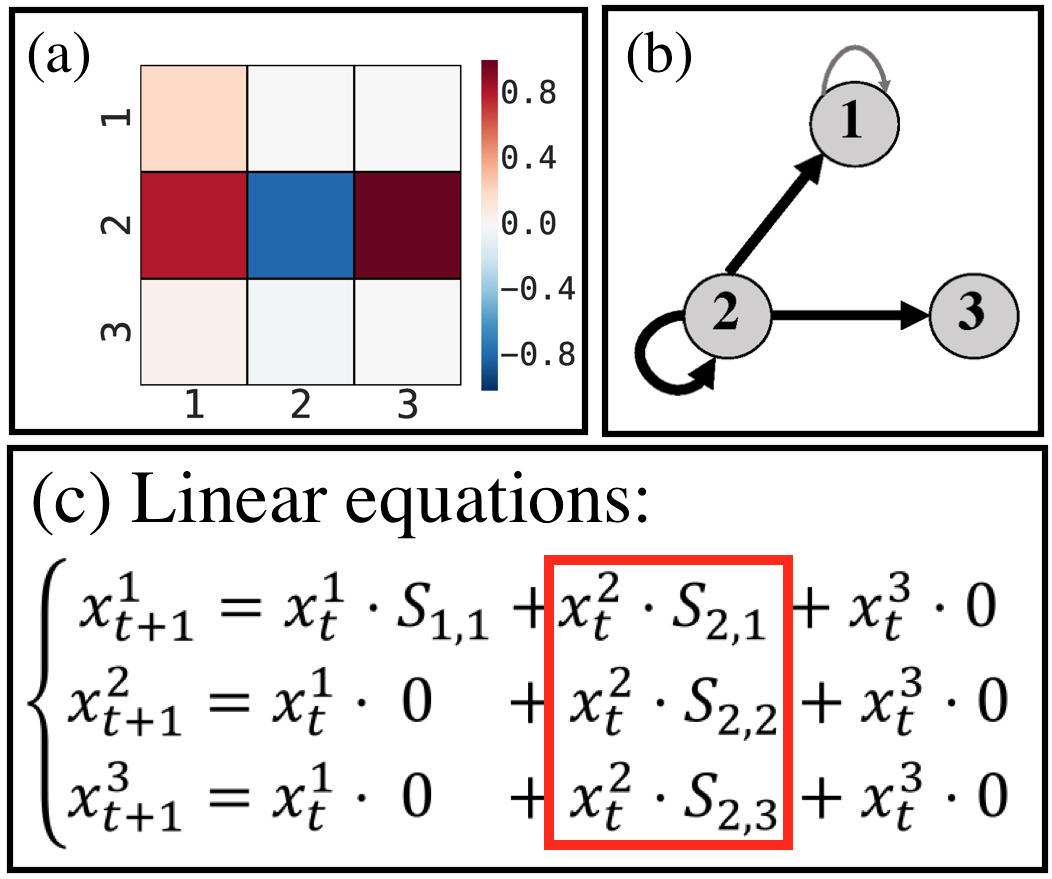}
\caption{Unimportant = row sparse. (a) Sentinel network $\mathbf{S}$; (b) the graph of $\mathbf{S}$; (c) the equations of a linear dynamical system, where the component $2$ dominates the system. Unimportant component $3$ is associated with a sparse row.}
\label{fig:basic}
\end{figure}

Our basic idea is intuitive: \emph{In a dynamical system, the components having little influence on others are unimportant for predicting others' states, while those exerting a heavy influence on others dominate the system dynamics and should be selected as sentinels.}
In terms of the sentinel network $\mathbf{S}$, one can determine whether a component is important or not by inferring row sparsity. 
That is, unimportant components are associated with sparse rows in $\mathbf{S}$, in which zeros are much more than non-zeros; on the other hand, important ones are associated with non-sparse rows. 
Figure\ref{fig:basic} shows an illustration by taking a linear dynamical system as an example. 

Based on this idea, we propose a novel index, the $\bm{\gamma}$ value, to measure components' importance in predicting the epidemic dynamics: a component is important if it is important in both prior and posterior structures of a sentinel network.
Specifically, the $\bm{\gamma}$ value is defined as the data-dependent hyper-parameter of the prior of the sentinel network, and also that reflecting the profiles of the posterior of the sentinel network:
\begin{equation}
\footnotesize
\footnotesize
\bm{\gamma}_i =((\bm{\mu}_{\bm{s}}^i)^{\rm T}  \bm{\mu}_{\bm{s}}^i + {\rm Tr}[ \bm{\Sigma}_{\bm{s}}^i ]) {N^{-1}},
\label{update_gamma}
\end{equation}
where $\bm{\gamma}_i$ is the $\bm{\gamma}$ value of component $i$.
In the following, we elaborate this importance measure from the perspectives of prior and posterior.

\subsubsection{3.2.1 ~ Prior perspective}
From the basic idea, a sentinel network is desired to have a row sparse structure. Thus, we adopt a zero-mean multivariate Gaussian prior for each row:
\begin{equation}
\footnotesize
p(\mathbf{S}_{i,\cdot}| \bm{\gamma}_i)\sim \mathcal{N}(\mathbf{0},\bm{\gamma}_i \mathbf{I}_N), ~ i = 1,\dots,N
\label{equ:prior_matrix}
\end{equation}
where vector $\mathbf{S}_{i,\cdot}$ $\in$ $\mathbb{R}^{N}$ denotes the $i$th row of the sentinel network, and $\mathbf{I}_N $ $\in$ $\mathbb{R}^{N \times N}$ is an identity matrix.
By doing so, $\bm{\gamma}_i$ controls the diversity of row $i$ from a zero vector.

For conciseness, we vectorize matrix $\mathbf{S}$, i.e., let $\bm{s} =$ vec($\mathbf{S}^\mathbf{T}$), where operator vec($\cdot$) denotes the vectorization of the input matrix by stacking its columns into a column vector.
By doing so, vector $\bm{s}$ $\in$ $\mathbb{R}^{N \times N}$ consists of $N$ groups of length $N$, where each group is associated with a row in $\mathbf{S}$.
Now, the row sparse structure of $\mathbf{S}$ is equal to the group sparse structure of $\bm{s}$.
In terms of the prior on $\mathbf{S}$ (Eq. \ref{equ:prior_matrix}), the prior over $\bm{s}$ is
\begin{equation}
\footnotesize
p(\bm{s}| \bm{\gamma}) \sim \mathcal{N}(\mathbf{0}, \bm{\Sigma}_0)
\label{equ:prior_vector},
\end{equation}
where vector $\bm{\gamma}$= $(\gamma_1,\cdots,\gamma_N)^{\rm T}$ and the covariance matrix $\bm{\Sigma}_0$ $\in$ $\mathbb{R}^{N^2 \times N^2}$ is a diagonal matrix:
\begin{equation}
\footnotesize 
\bm{\Sigma}_0 = \left[
\begin{smallmatrix}
\mathbf{\gamma}_1 \mathbf{I}_N    &          &       \\
        & \ddots &      \\
       &             &    \mathbf{\gamma}_N \mathbf{I}_N     \\
\end{smallmatrix}
\right].
\label{equ:sigma}
\end{equation}

As mentioned before, the links sent from a component reflect its effect on the system dynamics. 
Now, the links sent from $i$ in $\mathbf{S}$ (i.e., the entries of group $i$ in $\bm{s}$) are tied together and controlled by a common data-dependent hyper-parameter $\bm{\gamma}_i$.
This hyper-parameter is a type of automatic relevance determination (ARD) mechanism \cite{mackay1995probable}:
when $\bm{\gamma}_i$ is small, the group $i$ in $\bm{s}$ is sparse and vice versa;
when the group $i$ is sparse, the links sent from $i$ are very weak in $\mathbf{S}$.
That is, component $i$ is unimportant and can be pruned out without losing much in prediction accuracy.

\subsubsection{3.2.2 ~ Posterior perspective}
The $\bm{\gamma}$ value also reflects the profile of the posterior of the sentinel network.
We model the sentinel network for two kinds of dynamical systems widely used to characterize diffusion phenomena in the real world: a linear continuous system and a logistical discrete system.

\textbf{\emph{Likelihood of linear system.}}
Starting from the linear continuous system, we first give the likelihood function and illustrate the pre-processing of dynamics data.
The system function of a linear continuous system is
\begin{equation}
\footnotesize
\mathbf{Y}=\mathbf{X}\mathbf{S} + \mathbf{V},
\label{equ:diffusion_matrix}
\end{equation}
where $\mathbf{Y}$  =  $\mathbf{D}_{2:T+1}$ $\in$ $\mathbb{R}^{T \times N}$ and $\mathbf{X}$ = $ \mathbf{D}_{1:T}$$\in$ $\mathbb{R}^{T \times N}$ are both extracted from the dynamics data $\mathbf{D}$.
$\mathbf{Y}$ is the epidemic dynamics later than $\mathbf{X}$ one time-unit.
Specifically, $\mathbf{Y}_{t-1}$ = $\mathbf{X}_{t}$ = $\mathbf{D}_{t}$ and $\mathbf{V}$ is a Gaussian noise matrix.

For convenience, we further transform Eq. \ref{equ:diffusion_matrix} into a vector form, 
$\bm{y} = \mathbf{\Phi} \bm{s} + \mathbf{v}$,
where vector $\bm{y} $ = vec($\mathbf{Y}^{\rm T}$) $\in$ $\mathbb{R}^{TN \times 1}$, $\bm{s}$ = vec($\mathbf{S}^\mathbf{T}$) $\in$ $\mathbb{R}^{N^2 \times 1}$, and $\mathbf{v} $= vec($\mathbf{V}^{\rm T}$) $\in$ $\mathbb{R}^{TN \times 1}$.
The matrix $\bm{\Phi} =$ Kron($\mathbf{X}$, $\mathbf{I}_N$) $\in$ $\mathbb{R}^{TN \times N^2}$, where the operator Kron($\cdot$,$\cdot$) represents the Kronecker product of two input matrices.
Now, based on the Gaussian noise assumption, the likelihood of the linear continuous system can be given as
\begin{equation}
\footnotesize
p(\bm{y}|\bm{\Phi}, \bm{s}, \lambda) \sim \mathcal{N}(\bm{\Phi} \bm{s},\lambda \mathbf{I}),
\label{equ:likelihood_con}
\end{equation}
where $\lambda$ denotes the noise level and  $\mathbf{I}$ is an identity matrix.

\textbf{\emph{Likelihood of logistical system.}}
For the logistical discrete system, the entry in the dynamics data $\mathbf{D}_{t,i}$ is represented by a Boolean value, $0$ or $1$, indicating whether component $i$ is ``infected'' at time $t$.
After the same data pre-processing, we adopt a Bernoulli distribution over each entry of $\bm{y}$, i.e., $\bm{y}_n$:
\begin{equation}
\footnotesize
p(\bm{y}_n = 1| \bm{\Phi}_{n}, \bm{s} ) = \sigma[\bm{\Phi}_{n} \bm{s}], ~ n = 1,\dots,TN,
\label{ber_pro}
\end{equation}
where $\sigma[\bm{\Phi}_{n} \bm{s}] =1 / (1+e^{-\bm{\Phi}_{n} \bm{s}})$ denotes the sigmoid function,
and $\bm{\Phi}_{n}$ is the $n$th row of $\bm{\Phi}$.
Then, the likelihood of the dynamics can be written as
\begin{equation}
\footnotesize
p(\bm{y}|\bm{\Phi}, \bm{s} )=\prod \nolimits_{n=1}^{TN} \sigma[\bm{\Phi}_{n} \bm{s}]^{\bm{y}_n} (1- \sigma[\bm{\Phi}_{n}\bm{s}] )^{1- \bm{y}_n}.
\label{likelihood_nonlinear}
\end{equation}


Now, based on the aforementioned prior and likelihood we have the following conclusion about the posterior:

\textbf{Theorem 1.}
\emph{For both the linear system and logistical system, the posteriors of the sentinel network are a Gaussian or an approximate Gaussian:
$p(\bm{s}|\bm{y},  \bm{\Phi},\Theta) \sim \mathcal{N}(\bm{\mu}_{\bm{s}} , \bm{\Sigma}_{\bm{s}})$}:

\emph{(1) Linear system: hyper-parameters set $\Theta=\{  \bm{\gamma}, \lambda \}$, }
\begin{equation}
\footnotesize \nonumber
\begin{aligned}
&  \bm{\mu}_{\bm{s}} = \lambda^{-1} \bm{\Sigma}_{\bm{s}} \bm{\Phi}^{\rm T} \bm{y}, ~
\bm{\Sigma}_{\bm{s}}^{-1} = \bm{\Sigma}_0^{-1} + \lambda^{-1} \bm{\Phi}^{\rm T} \bm{\Phi};
\end{aligned}
\end{equation}

\emph{(2) Logistical system: hyper-parameters set $\Theta=\{\bm{\gamma}, \bm{\xi} \}$,}
\begin{equation}
\footnotesize \nonumber
\begin{aligned}
&  \bm{\mu}_{\bm{s}} = 2^{-1} \bm{\Sigma}_{\bm{s}} \bm{\Phi}^{\rm T}(2\bm{y} - \mathbf{1}) ,~
\bm{\Sigma}_{\bm{s}}^{-1} = \bm{\Sigma}_0^{-1} + \bm{\Phi}^{\rm T} \bm{\pi} (\bm{\xi}) \bm{\Phi},
\end{aligned}
\end{equation}
\emph{where $\bm{\xi}$= $[\xi_1, \dots, \xi_{TN}]^{\rm T}$ denotes variational parameters, and $\bm{\pi} (\bm{\xi}) $ $\in$ $\mathbb{R}^{TN \times TN}$ is a diagonal matrix. Specifically, $\bm{\pi} (\bm{\xi})_{n,n}$ = $- \frac{1}{2\xi_n}(\sigma [\xi_n] - \frac{1}{2})$.}
\begin{proof}
See the Supporting Information.
\end{proof}

\textbf{\emph{Estimation of hyper-parameters.}}
Theorem 1 gives the posteriors of the two kinds of systems (i.e., the Gaussian mean $\bm{\mu_s}$ and covariance matrix $\bm{\Sigma_s}$).
In the following,  we study how to iteratively update the hyper-parameters in the posteriors, because they cannot be obtained in a closed form.
By treating the $\bm{s}$ as hidden variables, we can employ the EM method to estimate the hyper-parameters set $\Theta$.
When the integral in the Q function can be analytically solved (in the linear system), the EM obtains a well-formed solution to estimating the hyper-parameters;
otherwise, we apply variational EM to estimate the hyper-parameters by optimizing an approximate low bound of the posterior  (in the logistical system).
Furthermore, the EM method can guarantee the convergence of the estimation.

Here, we only give the learning rules for the hyper-parameters since the limited space (derivation details can be found in the Supporting Information).
In the linear continuous system, the learning rule for the noise parameter $\lambda$ is
\begin{equation}
\footnotesize 
\lambda \leftarrow (TN)^{-1} (\parallel \bm{y}-\mathbf{\Phi} \mathbf{\mu}_{\bm{s}}  \parallel_2^2 + {\rm Tr} [\bm{\Sigma_{\bm{s}}} \mathbf{\Phi}^{\rm T} \mathbf{\Phi} ]).
\label{update_lambda}
\end{equation}
In the logistical discrete system, we update the variational parameter vector $\bm{\xi}$  by
\begin{equation}
\footnotesize  
\xi_n \leftarrow \sqrt{ \bm{\Phi}_n (\Sigma_{\bm{s}} +\bm{\mu}_{\bm{s}} \bm{\mu}_{\bm{s}}^{\rm T} ) \bm{\Phi}_n ^{\rm T}}, ~ n = 1,\dots,TN.
\label{update_xi}
\end{equation}
It is extremely interesting that in both the linear and logistical systems the learning rule for the $\bm{\gamma}$ value is the same, 
\begin{equation}
\footnotesize
\bm{\gamma}_i \leftarrow ((\bm{\mu}_{\bm{s}}^i)^{\rm T}  \bm{\mu}_{\bm{s}}^i + {\rm Tr}[ \bm{\Sigma}_{\bm{s}}^i ])N^{-1}, ~  i=1,\dots,N,\label{update_gamma}
\end{equation}
where the vector $\bm{\mu}_{\bm{s}}^i$ $\in$ $\mathbb{R}^{N}$  and matrix $\bm{\Sigma}_{\bm{s}}^i$ $\in$ $\mathbb{R}^{N \times N}$ denote the $i$th group of $\bm{\mu}_{\bm{s}}$ and $\bm{\Sigma}_{\bm{s}}$, respectively.

Intuitively, there are two terms that contribute to the $\bm{\gamma}$ value according to its learning rule, Eq .\ref{update_gamma}.
The first term is the inner product term $(\bm{\mu}_{\bm{s}}^i)^{\rm T} \bm{\mu}_{\bm{s}}^i$, which denotes the sum of the squares of the mean weights of the overall links sent from node $i$ in the sentinel network. In other words, it characterizes the \emph{influence strength} of component $i$. 
The second term is the trace term ${\rm Tr}[\bm{\Sigma}_{\bm{s}}^i]$, which denotes the variance of the posterior estimation on links sent from node $i$; 
that is to say, it features the \emph{influence uncertainty }of component $i$. 
In summary, a larger $\bm{\gamma}$ value corresponds to a node that has many links with large and diverse influences on other nodes.

On the whole, the $\bm{\gamma}$ value is an index by which the importance of a component for predicting the epidemic dynamics of an entire system can be measured.
This index integrates the profiles of both the prior and posterior of a sentinel network.
For a trivial component in the system, its $\bm{\gamma}$ value will tend to be zero due to the ARD mechanism during Bayesian learning.
For an important component, whose $\bm{\gamma}$ value larger than zero, its $\bm{\gamma}$ value could indicate its monitoring priority.
A component with a larger $\bm{\gamma}$ value exerts a great influence on other components, and its state is important to monitor for making a prediction.

Based on the $\bm{\gamma}$ value, we propose a backward-selection algorithm called the SNMA, as shown in Algorithm \ref{algor}.
It starts with all $N$ components of interest and removes one component at a time until only $k$ components are left ($k$ is according the budgets). 
The component that is removed should be chosen as the one with the minimum $\bm{\gamma}$ value.
The form of backward-selection algorithm is theoretically guaranteed to pick a optimal subset of components if the system perturbation is small enough \cite{couvreur2000optimality}.
A trade-off between accuracy and budget is practically necessary: the more sentinels are selected, the more predictive accuracy is expected, while more cost is needed.

\begin{algorithm}[htb]
\small
\caption{SNMA}
\KwIn{epidemic dynamics $\mathbf{D}$, quantity of components of interest $N$, quantity of sentinels $k$;}  
\KwOut{posterior structure of sentinel network, i.e., mean vector $\bm{\mu}_{\bm{s}}$, covariance matrix  $\bm{\Sigma}_{\bm{s}}$;}  
Pre-processing: extract $\bm{y}$ and $\bm{\Phi}$ from $\mathbf{D}$; \\
Randomly initialize $\bm{\gamma}$, $\lambda$ (the linear) or $\bm{\xi}$ (the logistical); \\
$L \leftarrow N$; \\
\While{$L>k$}
{
\While{$\bm{\gamma}$ is not converged}   
{
~~ $//$ \emph{The optimization step} \\
update $ \bm{\mu}_{\bm{s}} $ and $\bm{\Sigma}_{\bm{s}}$ via Theorem 1; \\
update $\bm{\gamma}$, $\lambda$ (the linear) or $\bm{\gamma}$, $\bm{\xi}$ (the logistical) via Eq. \ref{update_gamma},\ref{update_lambda} and \ref{update_xi}; 
}
find the minimum entry $i$ in the vector $\bm{\gamma}$\\
~~ $//$ \emph{The selection step} \\
update $\bm{\Phi}$, $ \bm{\mu}_{\bm{s}} $, $\bm{\Sigma}_{\bm{s}}$, $\lambda$ $\bm{\gamma}$, $\bm{\xi}$ through pruning out the entries of $i$th group in them;\\
$L \leftarrow L-1$;
}
\label{algor}
\end{algorithm}

\subsection{3.3 ~ Sentinel prediction}

Once we have obtained the posterior structure of the sentinel network, the epidemic dynamics of the overall system, $\mathbf{D}$, can be predicted based on the surveillance data $\mathbf{D}^{\bm{s}}$.
Let $\mathbf{D}^{\bm{s}}_{*}$ be a new set of surveillance data, where only the values on $k$ sentinels' locations are kept and the rest are empty.
As mentioned above, we obtain $\bm{\Phi}_{*}^{\bm{s}}$ through the data pre-processing. 
Then, a predictive distribution over the following system states $\bm{y}_*$ is given by
\begin{equation}
\footnotesize
p(\bm{y}_{*} | \bm{\Phi}_{*}^{\bm{s}}, \bm{y}, \bm{\Phi}) = \int
p(\bm{y}_{*} | \bm{\Phi}_{*}^{\bm{s}}, \bm{s})
p(\bm{s} | \bm{y}, \bm{\Phi})
d\bm{s}. \label{equ:sen_prediction}
\end{equation}

\emph{\textbf{Linear continuous system.}}
In this case, the integral in Eq. \ref{equ:sen_prediction}  is a Gaussian convolution (refer to the proof of Theorem 1), whose analytical solution is a Gaussian. Then, we have
\begin{equation}
\footnotesize \nonumber
p(\bm{y}_{*} | \bm{\Phi}_{*}^{\bm{s}}, \bm{y}, \bm{\Phi})  \sim
\mathcal{N}(\mu_{y_{*}}, \sigma_{y_{*}}^{2} )
\end{equation}
with parameters $\mu_{y_{*}} =  \bm{\Phi}_{*}^{\bm{s}} \bm{\mu}_{\bm{s}}$, $\sigma_{\bm{y}_{*}}^{2} = \lambda + \bm{\Phi}_{*}^{\bm{s}} \bm{\Sigma_{\bm{s}}}  (\bm{\Phi}_{*}^{\bm{s}})^{\rm T}$.

\emph{\textbf{Logistical discrete system.}}
The predictive distribution of discrete data is a Bernoulli distribution.
By substituting the term of the posterior of $\bm{s}$ in Eq. \ref{equ:sen_prediction} for the variational approximation posterior given in Theorem 1, we have the following predictive distribution:
\begin{equation}
\footnotesize \nonumber
p(\bm{y}_{*n}=1 | \mathbf{\Phi}_{*}^{\bm{s}}, \bm{y}, \mathbf{\Phi}) \approx \int
p(\bm{y}_{*n}=1 | \mathbf{\Phi}_{*}^{\bm{s}}, \bm{s})
q(\bm{s} | \bm{y}, \mathbf{\Phi})
d\bm{s},
\end{equation}
where $n = 1 \cdots N$.

However, the integral in this approximation is a convolution of a logistical function with a Gaussian distribution, which cannot be analytically integrated. 
By introducing an error function, it can be approximated as a re-parameterized logistical function \cite{maragakis2008bayesian}:
\begin{equation}
\footnotesize 
\int p(\bm{y}_{*} \text{=}1 | \mathbf{\Phi}_{*}^{\bm{s}}, \bm{s})  q(\bm{s} | \bm{y}, \mathbf{\Phi}) d\bm{s} \approx
(1+e^{- \tau \mathbf{\Phi}_{*}^{\bm{s}} \bm{\mu_{\bm{s}}}})^{-1},
 \label{equ:prediction_discrete}
\end{equation}
where
$ \tau = (1+\frac{\pi}{8}  \mathbf{\Phi}_{*}^{\bm{s}} \bm{\Sigma_{\bm{s}}}  (\mathbf{\Phi}_{*}^{\bm{s}})^{\rm T} )^{-\frac{1}{2}}.$
A very small approximation error is guaranteed in theory \cite{maragakis2008bayesian}.

\subsection{3.4 ~ Scaling up}

In the SNMA, the most expensive operations are the matrix multiplication ($\bm{\Phi}^{\rm T} \bm{\Phi}$) and the matrix inverse ($\bm{\Sigma}_{\bm{s}}^{-1}$) for calculating the posterior parameters in Theorem 1. 
Considering the large size of $\bm{\Phi}$ $\in$ $\mathbb{R}^{TN \times N^2}$ and $\bm{\Sigma}_{\bm{s}}$ $\in$ $\mathbb{R}^{N^2 \times N^2}$, the time complexity of one iteration will be $O(N^5 \times \max(T,N))$, making it infeasible to handle real-world problems. 
We now propose two fast matrix operations to solve the scalability problem. 
The time complexity after scaling up is reduced to $O(N^2 \times \max(T,N))$, which is faster than the competitors in the experiments.

\begin{figure}[h]
\centering{}
\includegraphics[width=0.45 \textwidth]{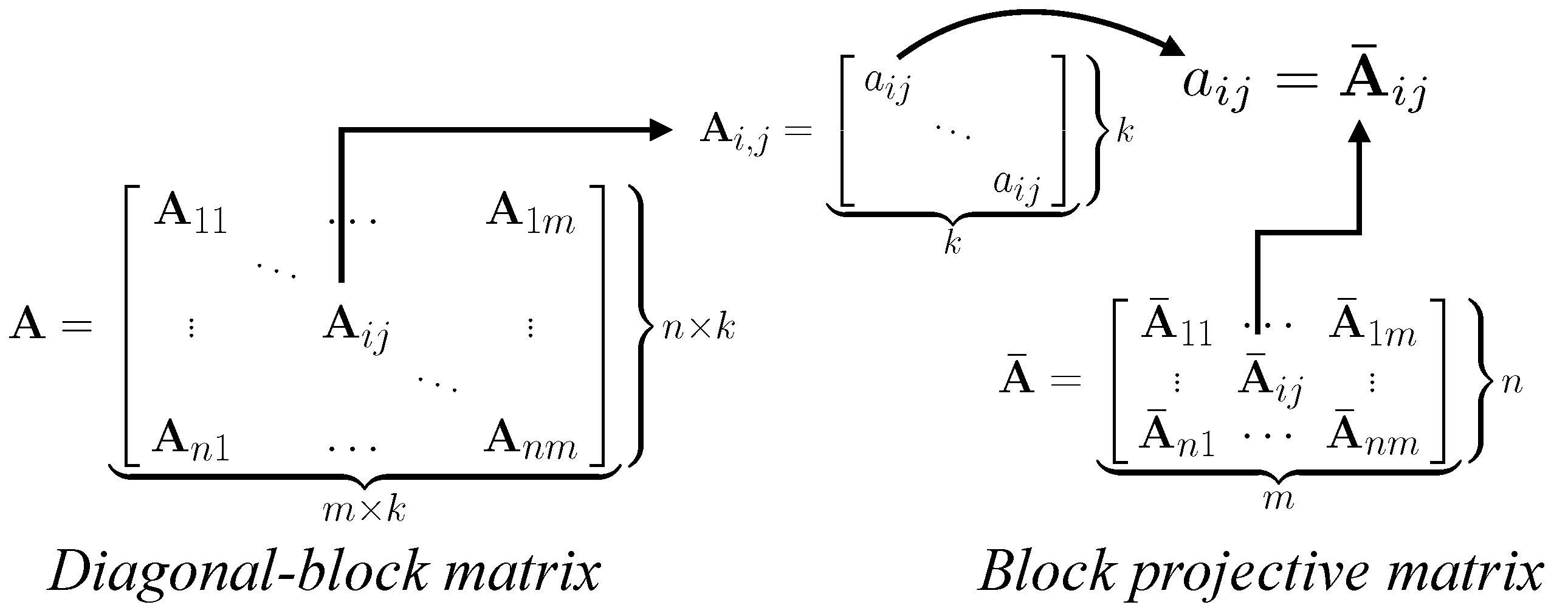}
\caption{Illustration of the relationship between a diagonal-block matrix and block projective matrix.}
\label{fig_matrix_trans}
\end{figure}

\noindent \textbf{Definition 1.}
\emph{ (Diagonal-block matrix). Matrix $\mathbf{A}$ $\in$ $\mathbb{R}^{nk \times mk}$ is called a diagonal-block matrix if it consists of $n$ by $m$ square sub-matrices, and each sub-matrix $\mathbf{A}_{i,j}$ $\in$ $\mathbb{R}^{k \times k}$ is a diagonal matrix with all diagonal entries having the same value $a_{ij}$, $i$=$1:n, j$=$1:m$.}

\noindent \textbf{Definition 2.}
\emph{ (Block projective matrix). Let $\mathbf{A}$ $\in$ $\mathbb{R}^{nk \times mk}$ be a diagonal-block matrix. Matrix $ \bar{\mathbf{A}}$ $\in$ $\mathbb{R}^{n \times m}$ is called the block projective matrix of $\mathbf{A}$ if each entry $\bar{\mathbf{A}}_{i,j} = a_{ij}$, $i$=$1:n, j$=$1:m$.}

The structure of the two matrices and their relationship is shown in Fig. \ref{fig_matrix_trans}.
The block projective matrix $ \bar{\mathbf{A}}$ is much smaller than its original diagonal-block matrix $\mathbf{A}$. 
In the meantime, it contains the overall distinct elements in $\mathbf{A}$.
Based on the definitions, we have the following theorems, which can significantly reduce the computational cost of the two kinds of operations.

\noindent \textbf{Theorem 2.}
\emph{ Let $\mathbf{A}, \mathbf{B}$ be two diagonal-block matrices with the same size of sub-matrix. The product $\mathbf{A}\mathbf{B}=\mathbf{C}$ is a diagonal-block matrix, and those block projective matrices satisfy $\bar{\mathbf{A}} \bar{\mathbf{B}} = \bar{\mathbf{C}}$.}
\begin{proof}
See the Supporting Information.
\end{proof}

\noindent \textbf{Theorem 3.}
\emph{ Let $\mathbf{A}$ be a square diagonal-block matrix. Its inverse matrix $\mathbf{A}^{-1}$ is also a diagonal-block matrix and satisfies 
$\bar{(\mathbf{A}^{-1})}$ = $(\bar{\mathbf{A}})^{-1}$.}
\begin{proof}
See the Supporting Information.
\end{proof}

In the SNMA,  $\bm{\Phi}$ and $\bm{\Sigma}_{\bm{s}}$ are two diagonal-block matrices.
When we alternatively calculate  the multiplication  ($\bm{\Phi}^{\rm T} \bm{\Phi}$) and the matrix inverse ($\bm{\Sigma}_{\bm{s}}^{-1}$)  on the  block projective matrices $\bm{\bar{\Phi}}$  $\in$ $\mathbb{R}^{T \times N}$ and $\bm{\bar{\Sigma}}_{\bm{s}}$  $\in$ $\mathbb{R}^{N \times N}$ via the two theorems, the time complexity of one iteration can be reduced by 3 orders of magnitude (i.e., $N^3$).
Note that this scaling-up technique not only solves the difficulty faced by our algorithm, but can also address other tasks involving diagonal-block matrices, such as the computational obstacle of a multiple measurement vector (MMV) model in compressed sensing \cite{zhang2011sparse}.
To process further large-scale data in practice, the SNMA can be readily parallelized by adopting the group testing strategy, which has been used for parallel feature selection \cite{zhou2014parallel}. 

\begin{figure*}[!hbt]
\centering
\begin{tabular}{cccc}
\includegraphics[width=0.23\textwidth]{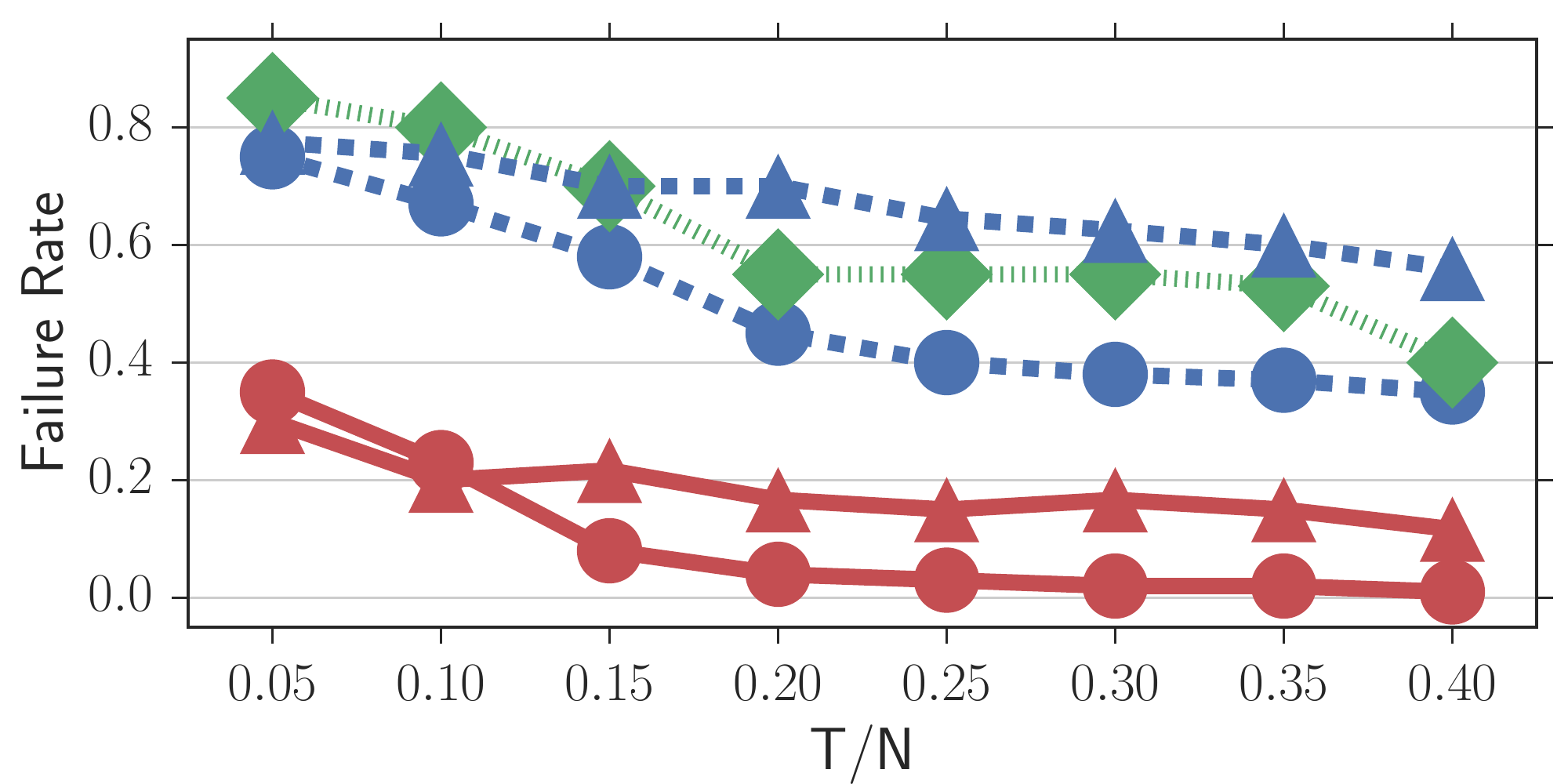} &
\includegraphics[width=0.23\textwidth]{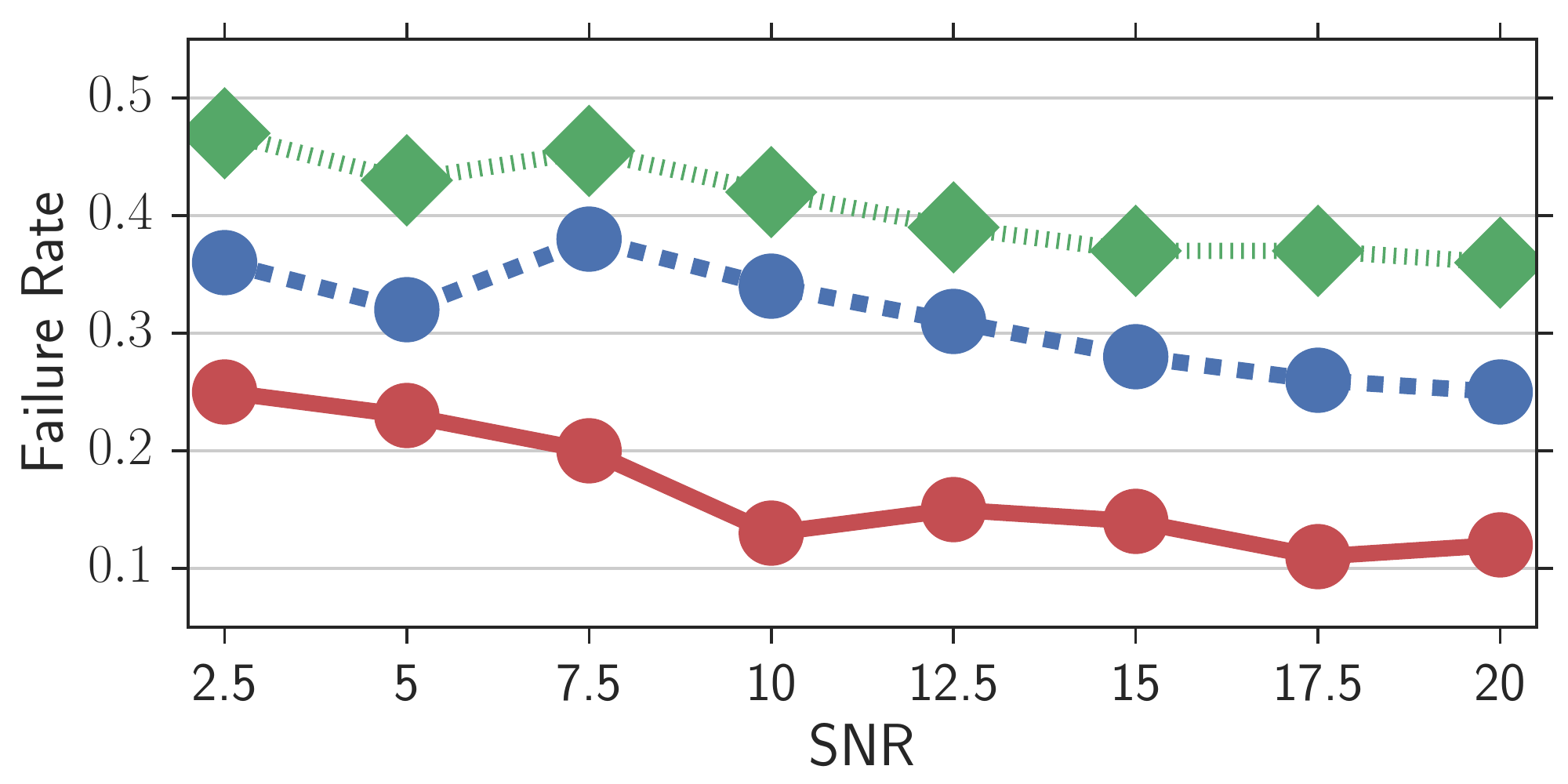} &
\includegraphics[width=0.23\textwidth]{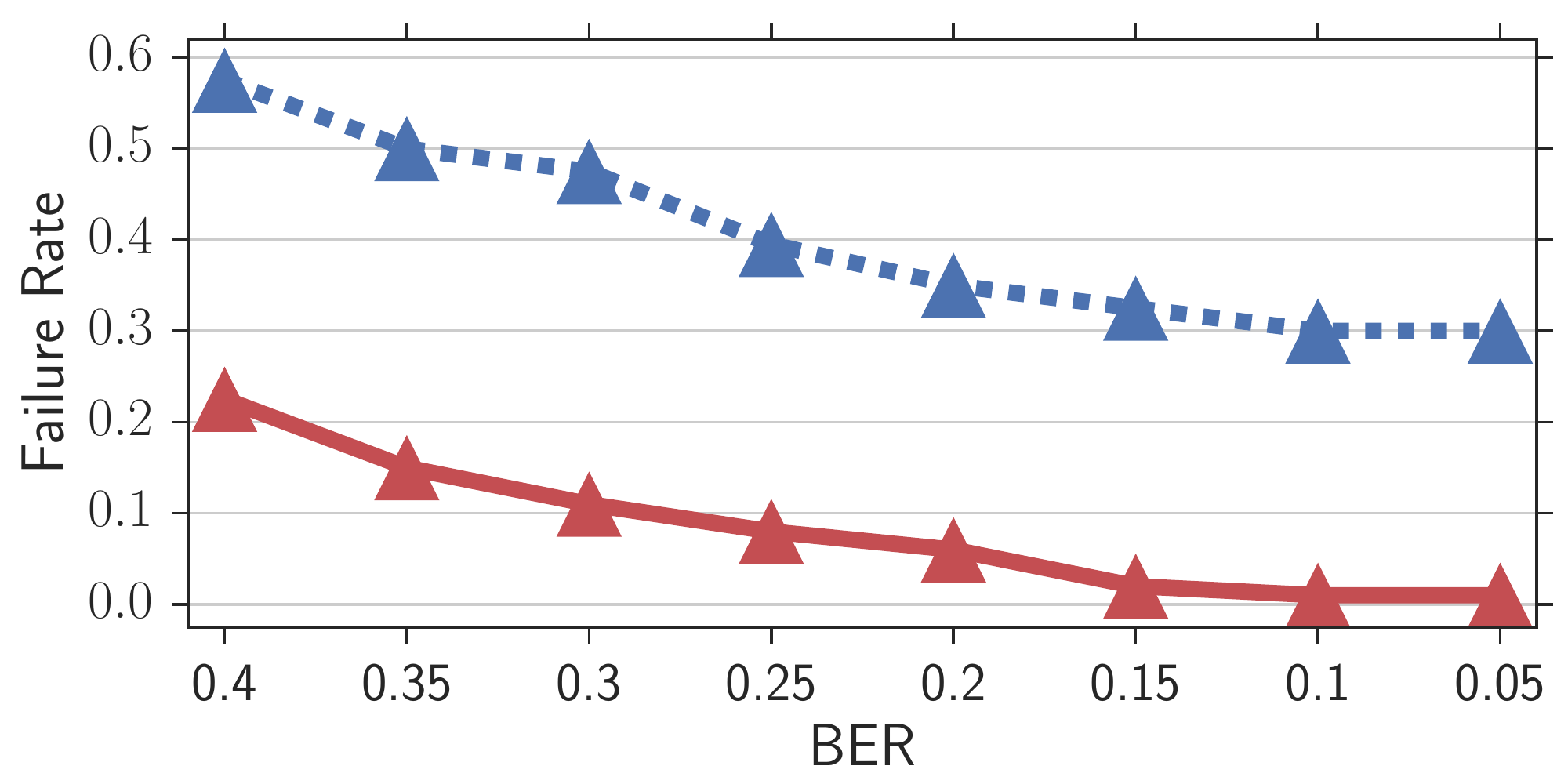} &
\includegraphics[width=0.23\textwidth]{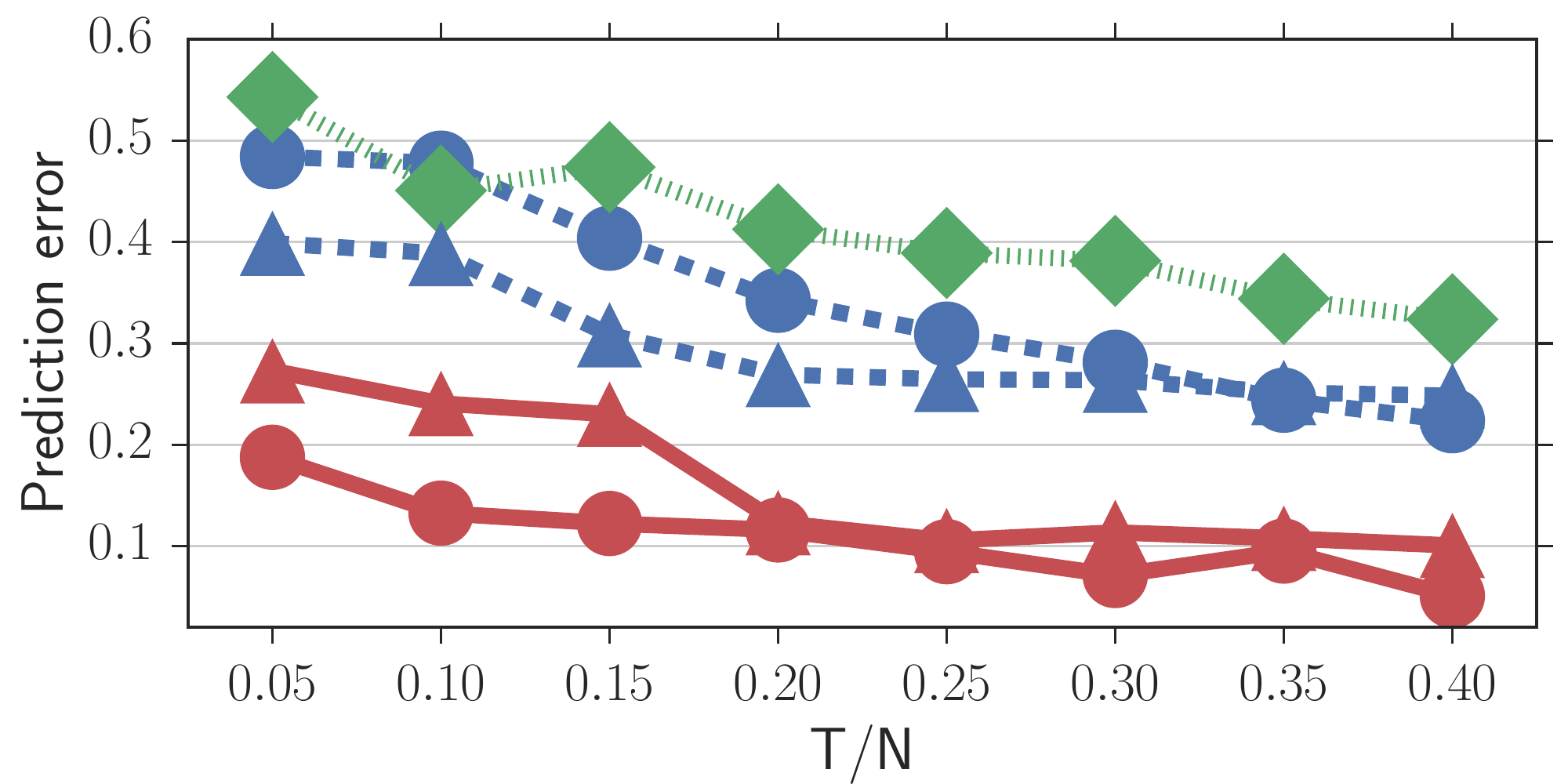}   \\
(a) & (b) & (c)& (d) \\
\includegraphics[width=0.23\textwidth]{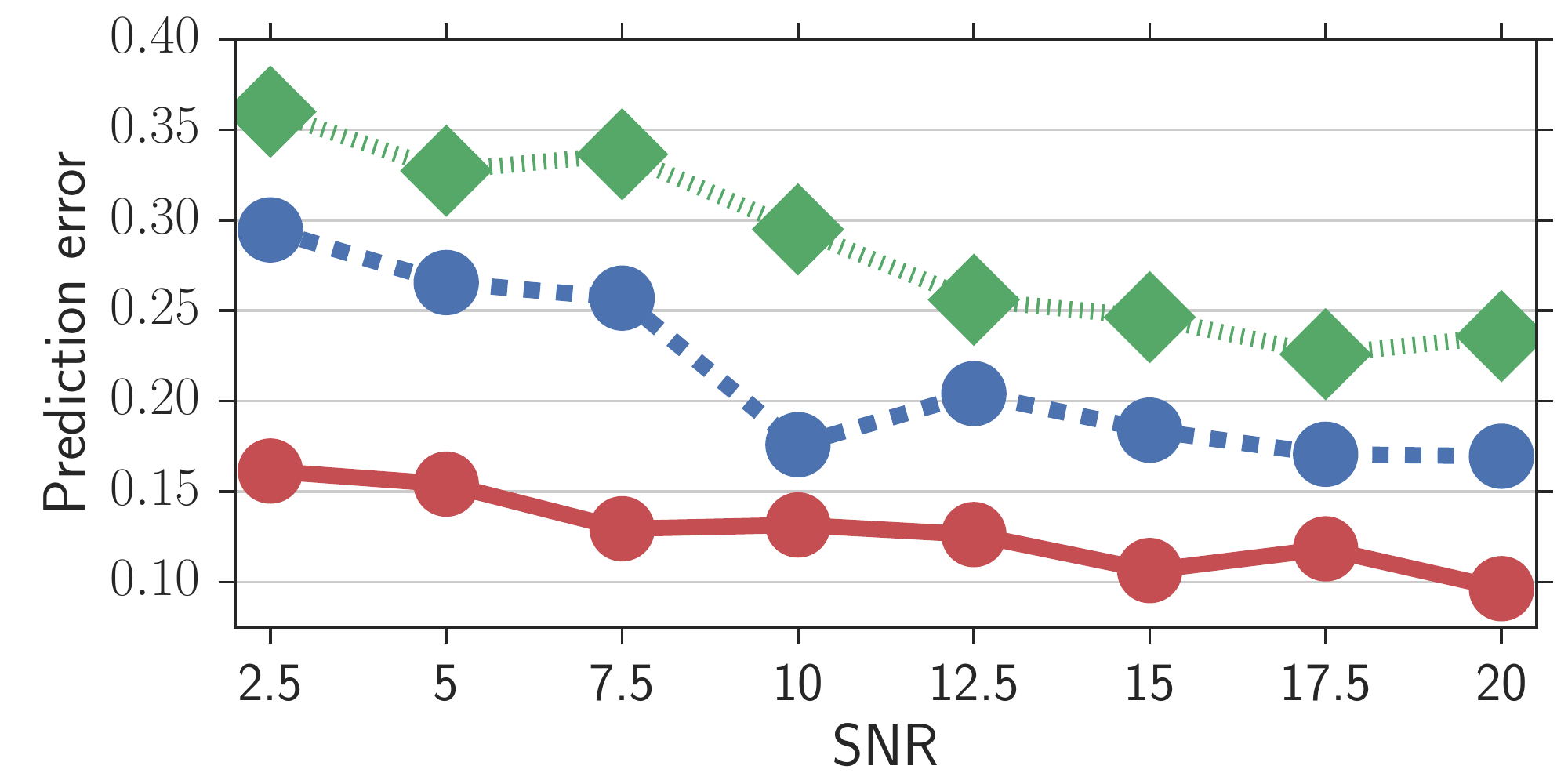} &
\includegraphics[width=0.23\textwidth]{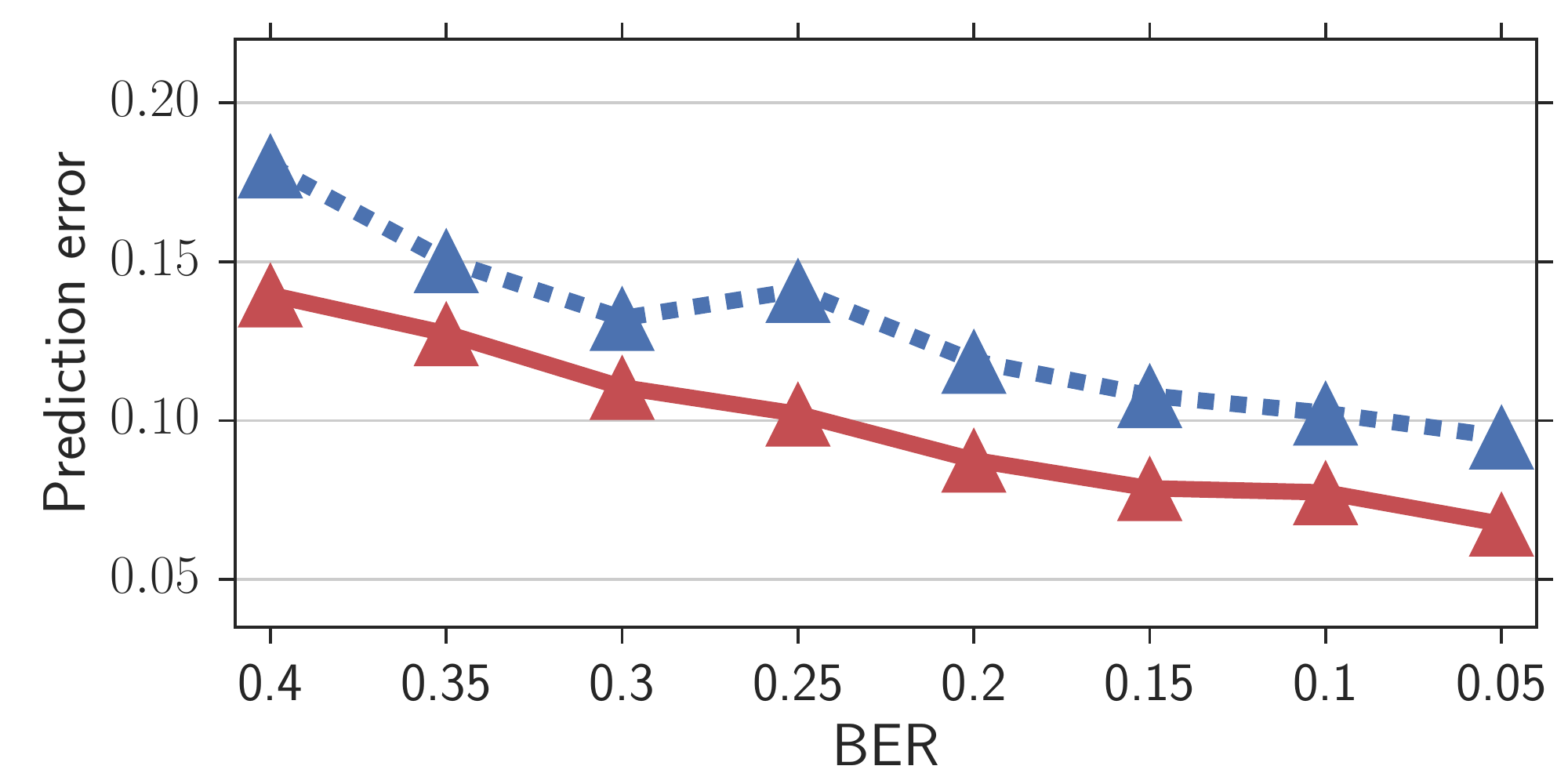} &
\includegraphics[width=0.23\textwidth]{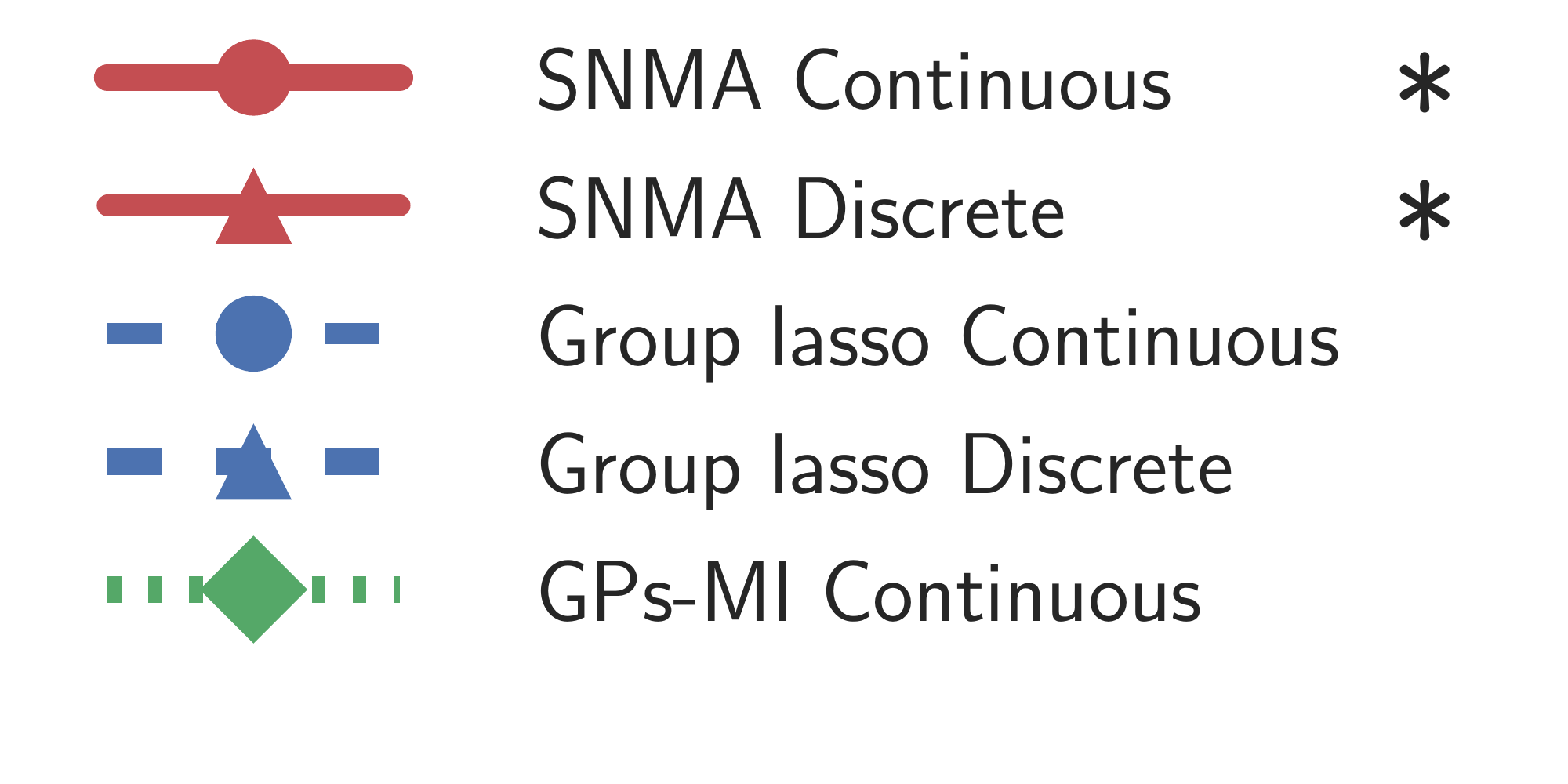} &
\includegraphics[width=0.23\textwidth]{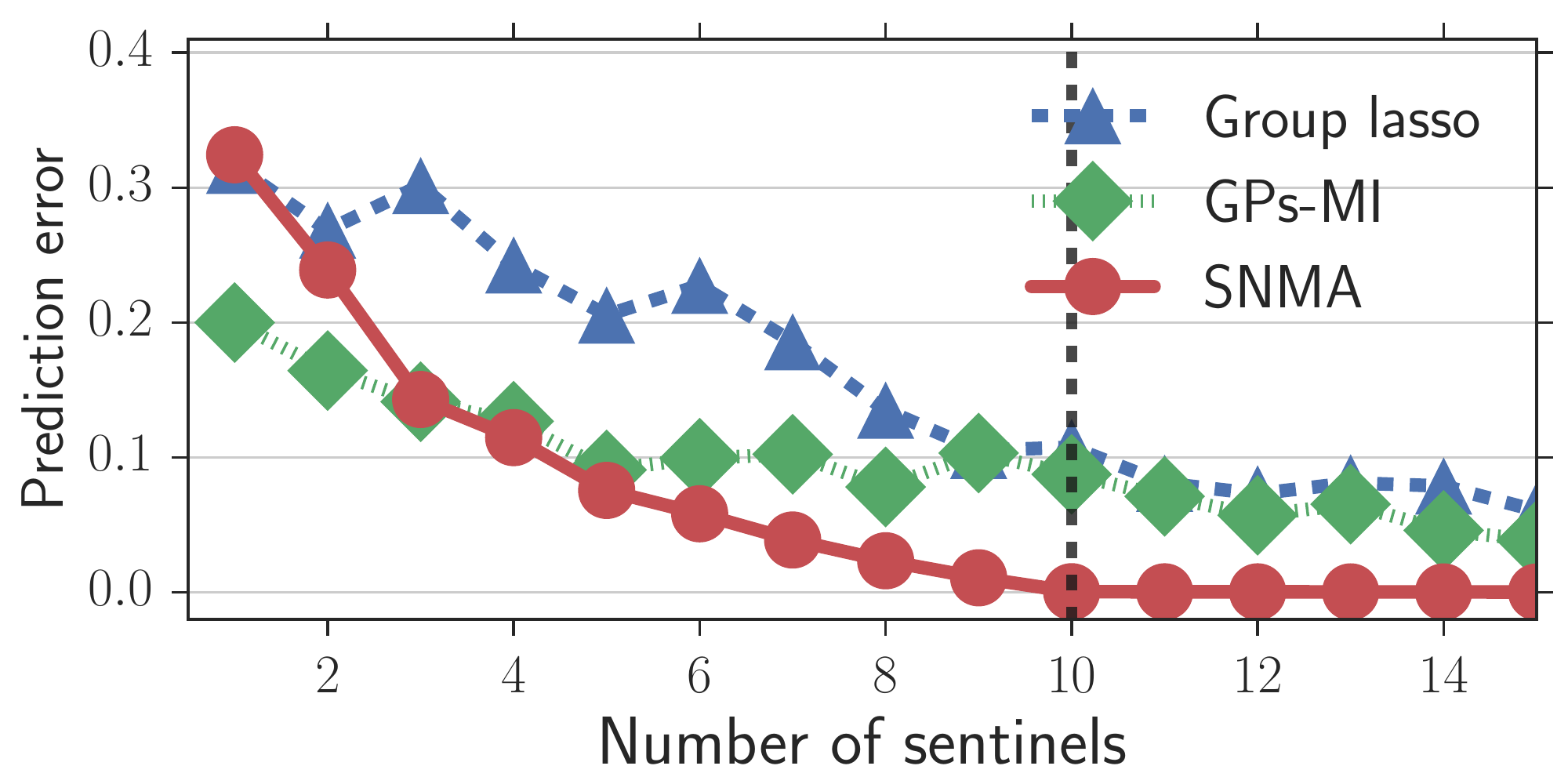}   \\
(e) & (f) & The legend of (a-f)  & (g)
\end{tabular}
\caption{Results of experiments on synthetic data. ($*$ denotes our methods).}
\label{fig:synt_result}
\end{figure*}

\subsection{3.5 ~ Embedding non-linear dynamical models}

The proposed framework is flexible and can be readily extended to various dynamical systems by mean of embedding basic functions, as long as the dynamical system functions can be represented as the combinations of basic functions. 
Each basic function, $\psi_i(x), i \in [1,\cdots,m]$, can be an arbitrary function, e.g., polynomial, $\psi_i(x)$= $x^2+x$; trigonometric, $\psi_i(x)$= $\sin (x)$; or others.

We illustrate the embedding technology based on the linear system Eq. \ref{equ:diffusion_matrix} as an example, where $\mathbf{X}_{t,i}$ denotes the state of component $i$ at time $t$. 
Let vector $\Psi(\mathbf{X}_{t,i})$ = $[\psi_1(\mathbf{X}_{t,i}),\cdots, \psi_m(\mathbf{X}_{t,i})]$ $\in$ $\mathbb{R}^{m}$ be the mapping of $\mathbf{X}_{t,i}$ through $m$ basic functions. 
Let vector $\Psi(\mathbf{X}_t)$ = $[\Psi(\mathbf{X}_{t,1}),\cdots,\Psi(\mathbf{X}_{t,N}) ]$ $\in$ $\mathbb{R}^{mN}$ denote the mapping of all components at time $t$, 
and matrix $\Psi(\mathbf{X})$ = $[\Psi(\mathbf{X}_{1});\cdots;\Psi(\mathbf{X}_{T}) ]^{\rm T}$ $\in$ $\mathbb{R}^{ T \times  mN}$ denote the mapping of all components during the time window.
Then, the system function Eq.\ref{equ:diffusion_matrix} can be extended as $\mathbf{Y} =   \Psi(\mathbf{X}) \mathbf{\dot{S}} + \mathbf{V}$,
\footnotesize
where the augmented sentinel network $\mathbf{\dot{S}}$ $\in$ $\mathbb{R}^{mN \times N}$ characterizes the interactions from the $mN$ mapping to the $N$ components. 
This extended equation can also represent non-linear dynamical systems if non-linear basic functions are adopted, such as the SIR model, a well-studied and widely adopted  disease spread model. 
Moreover, the basic functions of the SIR model can be constructed based on its reproduction matrix form \cite{wallinga2010optimizing}.

To integrate the above extended system function into the current SNMA, just let $\bm{\Phi} =$ Kron($\Psi({\mathbf{X}})$, $\mathbf{I}_N$) $\in$ $\mathbb{R}^{TN \times mN^2}$ and $\bm{s} =$ vec($\mathbf{\dot{S}})^\mathbf{T}$.
Meanwhile, the group size needs be changed from $N$ to $mN$.
The rest of the steps are the same as the SNMA shows in Algorithm \ref{algor}.
For the logistical system, the extended process is analogous.

\section{4 ~ Validations}

\emph{Comparative Study.}
We validate the framework on both synthetic and three real-world data.
The two most related methods are selected as competitors: group Lasso and GPs-MI.
Group Lasso  \cite{meier2008group} is a typical method for group sparse learning, which is similar to that addressed by our proposed group sparse Bayesian learning algorithm.
To achieve the aim of active surveillance, we use the proposed framework and replace the group sparse Bayesian learning with group lasso. 
 
GPs-MI is a popular sensor placement method \cite{krause2008near,hoang2014nonmyopic}, which is similar to the task of active surveillance addressed by our work. 
It outperforms the placement methods based on experiment design, such as A-, D-, and E-optimal designs.
As Gaussian processes cannot directly work on discrete data, GPs-MI is only applied on experiments with continuous data.

\emph{Evaluation.}
Two criteria are adopted to evaluate the performance of the methods.
Failure rate is used to measure whether the a method can discover the sentinels, i.e., the problem of sentinel identification. 
Failure rate is the percentage of wrong sentinels' locations given by a method, $1- | \mathbf{\Gamma} \cap \hat{\mathbf{\Gamma}} |/ |\mathbf{\Gamma}|$, where $\mathbf{\Gamma}$ is the set of the ground truth sentinel locations in, and $\hat{\mathbf{\Gamma}}$ is the set discovered by, a method.

We use \emph{root-mean-square error} (RMSE)  to quantify the prediction error,  i.e., the problem of sentinel prediction.
Specifically, RMSE=$\frac{ ||\mathbf{Y} - \hat{\mathbf{Y}}||_2 }{ TN}$, where $\hat{\mathbf{Y}}$ is the predicted epidemic dynamics based on the surveillance data.

\subsection{4.1 ~ Validations on synthetic data}

\emph{Synthetic Data Generation.}
We generate synthetic data by imagining diffusion processes taking place in a linear continuous system or logistical discrete system. 
There are three steps in total: 
(1) Random generation of a ground truth $\bm{\gamma}$-value vector with $500$ entries, where $100$ entries are sampled from $\mathcal{N}(0, 10)$ and the other $400$ from $\mathcal{N}(0, 0.1)$, i.e.,  $100$ sentinels and $400$ trivial components;
(2) Random sampling of a ground truth sentinel network $\mathbf{S}$ via the prior Eq.\ref{equ:prior_matrix} based on the ground truth $\bm{\gamma}$ value; 
(3) Based on the $\mathbf{S}$, simulation of the epidemic dynamics via the linear system Eq. \ref{equ:diffusion_matrix} or the logistical system Eq. \ref{ber_pro}  embedding a quadratic basic function $\phi(x) = x^2 + x$. 

\emph{Environment setting.}
The comparisons are conducted under various data volumes and noise levels.
We use the value $T/N$ to denote the ratio of data volume to the number of parameters to estimate.
$\emph{Signal-to-noise ratio}$ (SRN) and $\emph{bit error rate}$ (BER) are adopted to denote noise levels in the linear system and logistical system, respectively.

We adopt a $5$-fold cross-validation strategy in experiments on synthetic data.
We firstly identify $100$ sentinels from the training data via the three methods.
For each method, the average failure rate of sentinel identification is shown in Figs. \ref{fig:synt_result} (a-c).
Then, we evaluate the performance of sentinel prediction by feeding the surveillance data (collected on the $100$ discovered sentinels) to the corresponding prediction model of each method, such as Eq. \ref{equ:sen_prediction} for the proposed method.
The average prediction error is shown in Figs.\ref{fig:synt_result}(d-f). 
The results show that  the proposed methods are superior to GPs-MI and group lasso on both failure rate and prediction error.

We evaluate the trade-off between the prediction accuracy and surveillance cost in the linear system by setting different number of sentinels as shown in Fig.\ref{fig:synt_result}(g).
In this experiment, there are only $10$ ground truth sentinels in $\mathbf{S}$. 
We give the number of sentinels $k$ ($x$-axis), and then evaluate the prediction error of each method ($y$ axis).
Intuitively, the more sentinels that are selected, the better the accuracy that can be obtained. 
However, as indicated, the improvement in prediction becomes negligible when $k$ is over $10$. 
Note that GPs-MI shows a better performance only when $k$ is very small.

\begin{figure}[h]
\centering{}
\includegraphics[width=0.325 \textwidth]{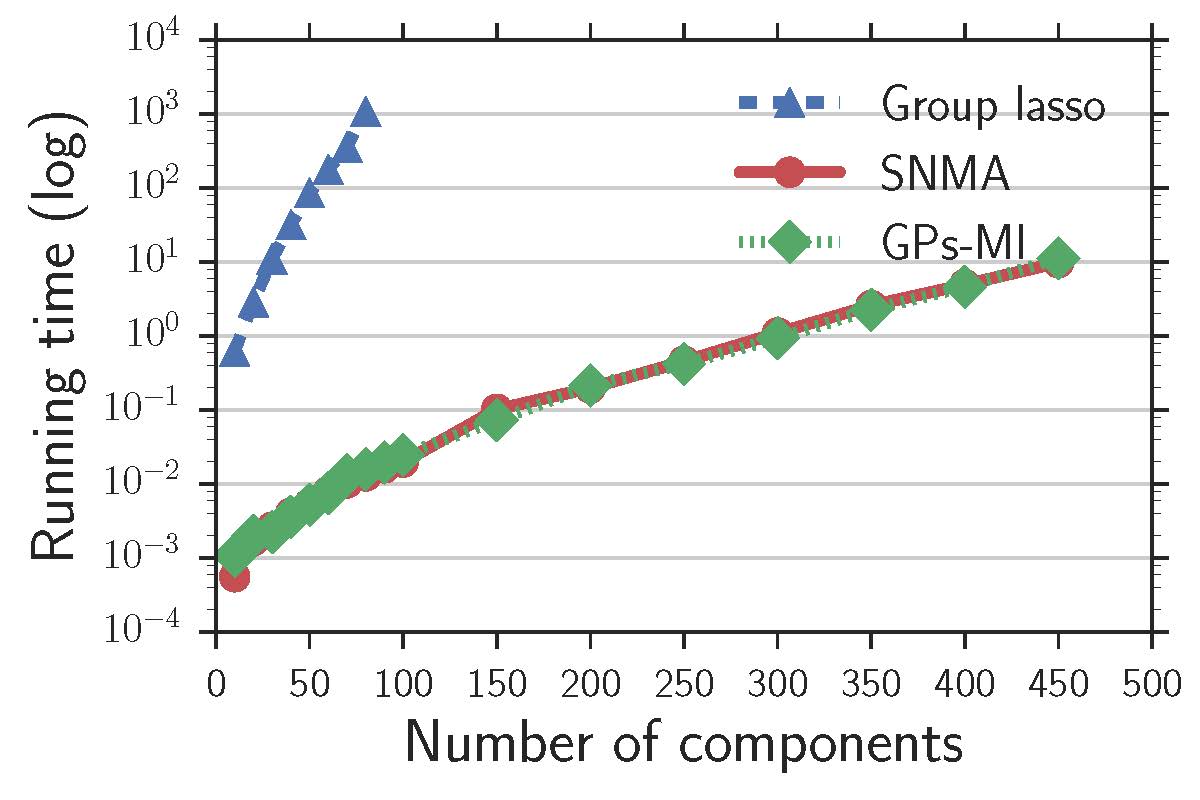}
\caption{Comparison on running time. $y$-axis is the time (seconds) on a log scale. $x$-axis denotes the size of a system.}
\label{fig:running_time}
\end{figure}

\begin{figure*}[h]
\centering
\begin{tabular}{cccc}
\includegraphics[width=0.315\textwidth]{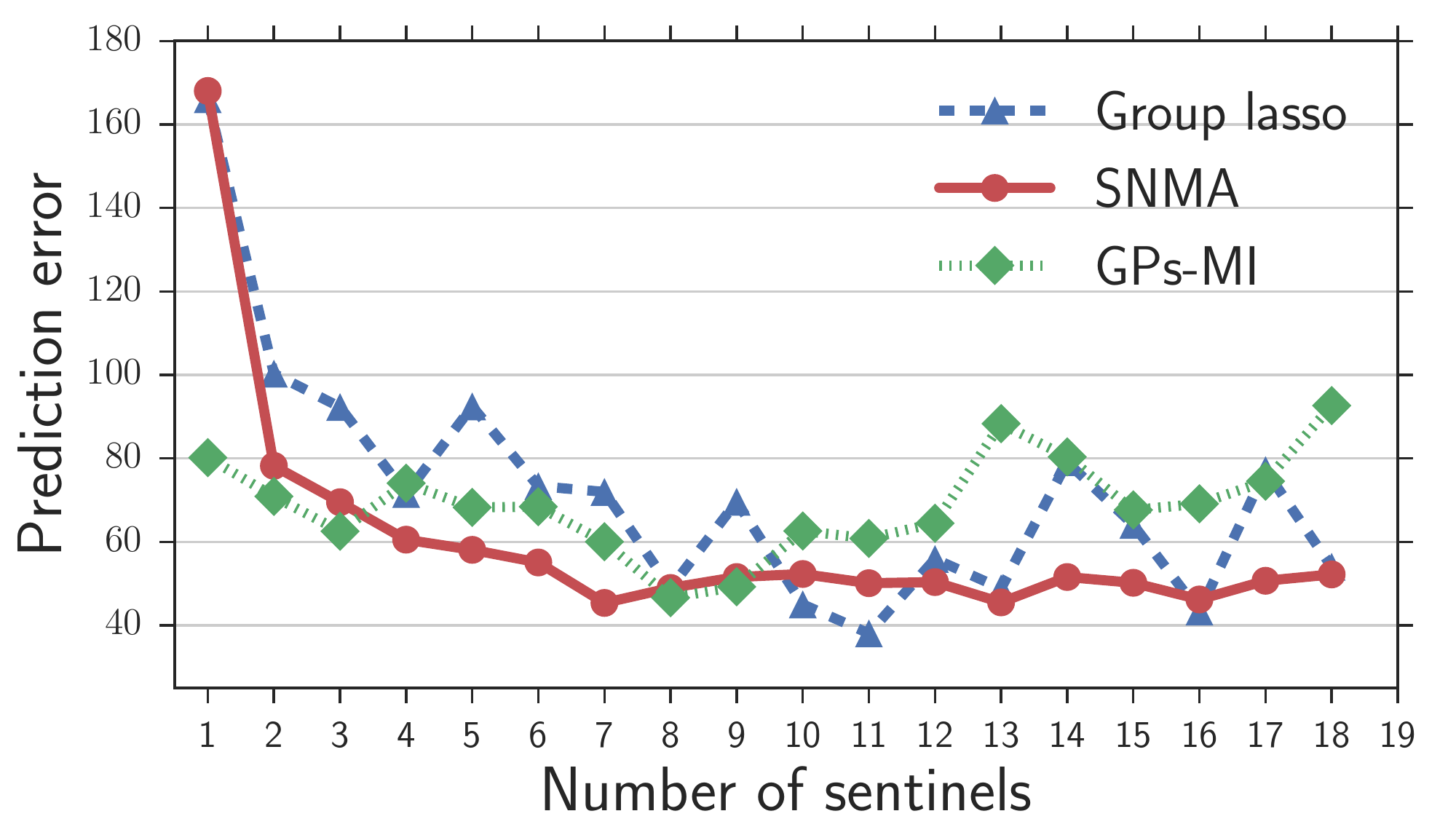} &
\includegraphics[width=0.315\textwidth]{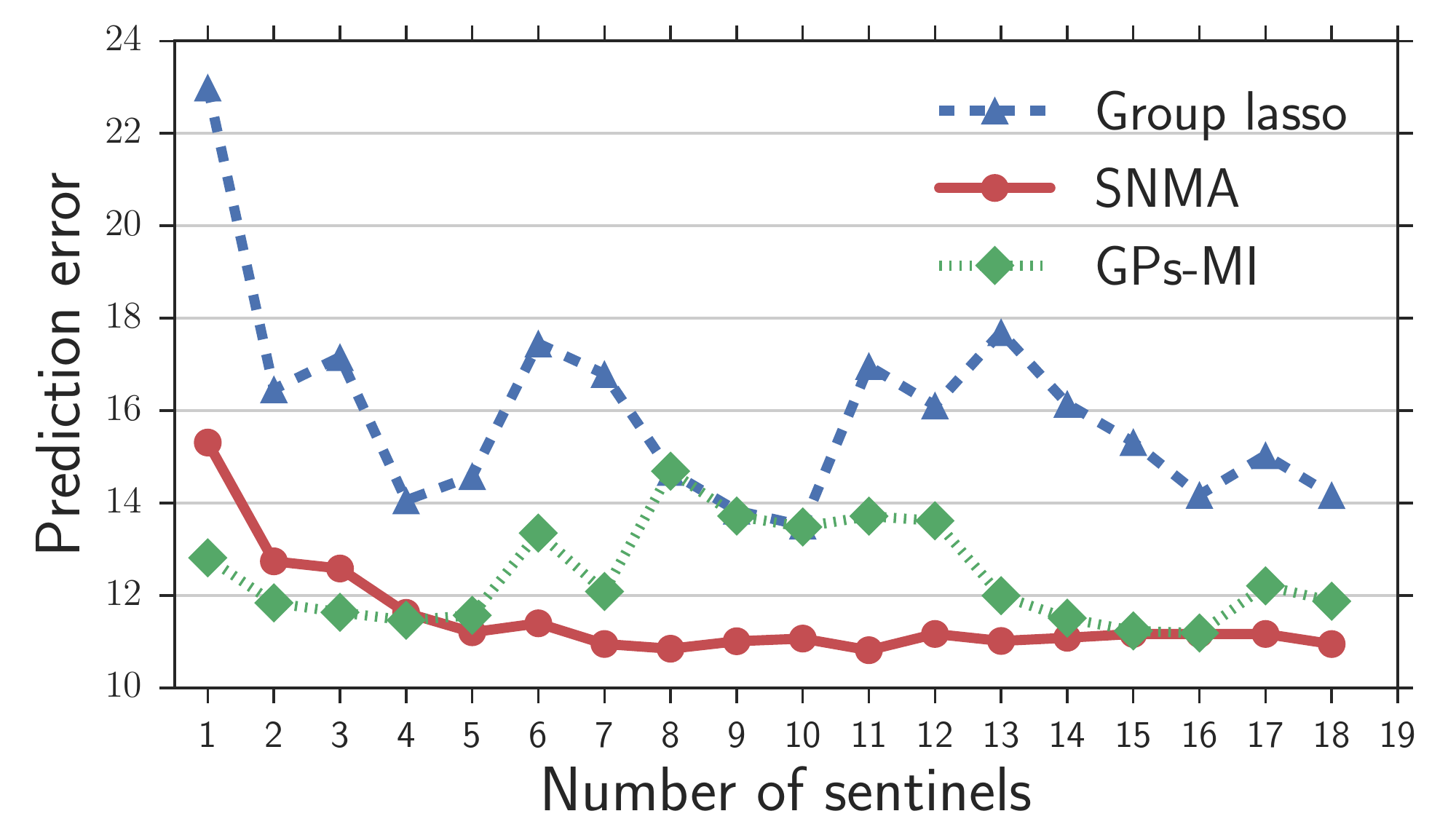} &
\includegraphics[width=0.315\textwidth]{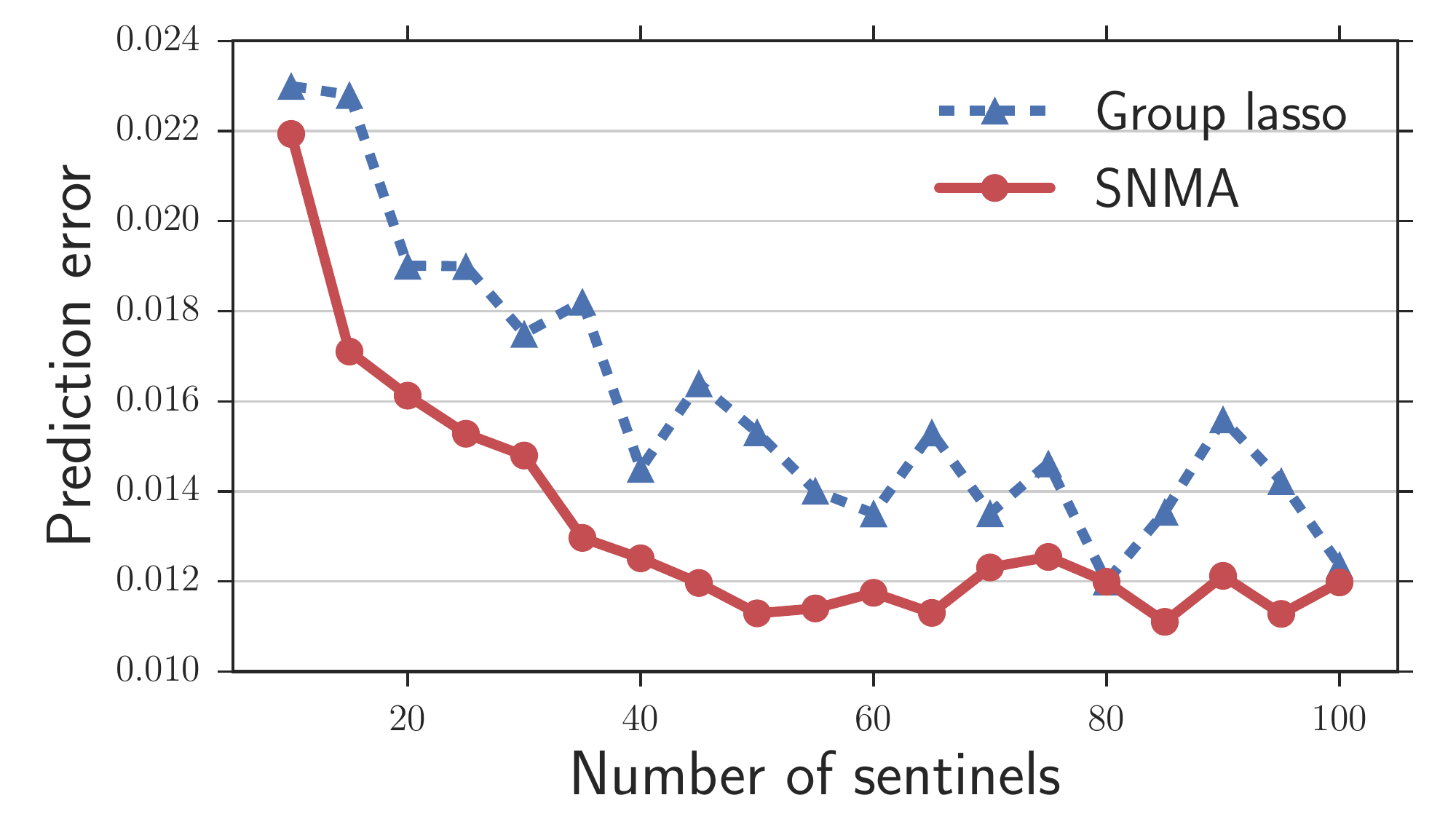} \\
(a) & (b) & (c)
\end{tabular}
\caption{Comparisons on the real-world epidemic dynamics. $x$-axis is the number of selected sentinels and $y$-axis denotes the prediction error (RMSE). (a) 2009 Hong Kong H1N1 flu pandemic.  (b) 2005-2009 Tengchong malaria outbreak. (c) The dynamics of hot words cascading in Baidu Tieba.}
\label{fig:sentinel_pre}
\end{figure*}

\begin{figure*}[h]
\centering
\begin{tabular}{cccc}
\includegraphics[width=0.31\textwidth]{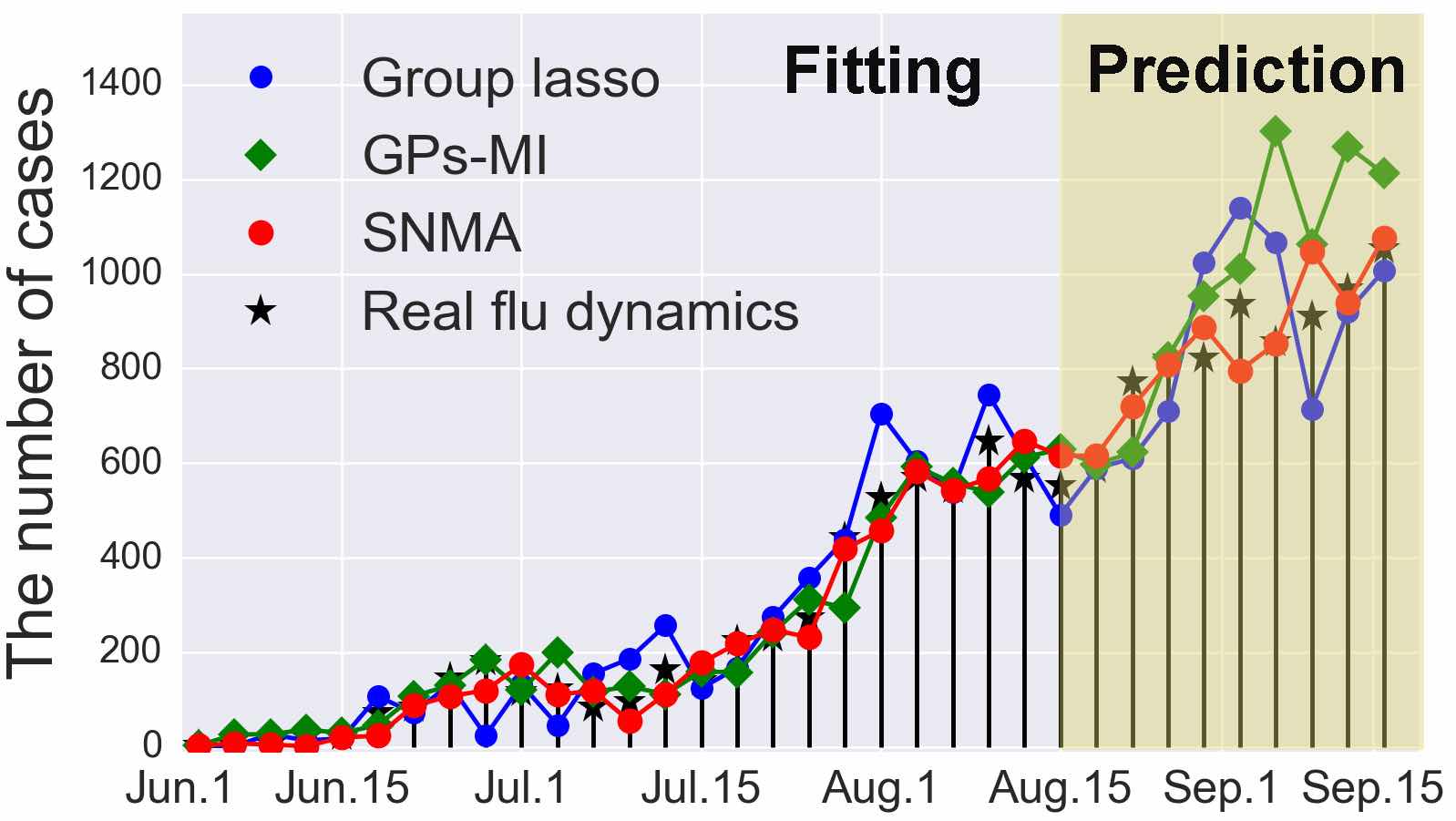} &
\includegraphics[width=0.31\textwidth]{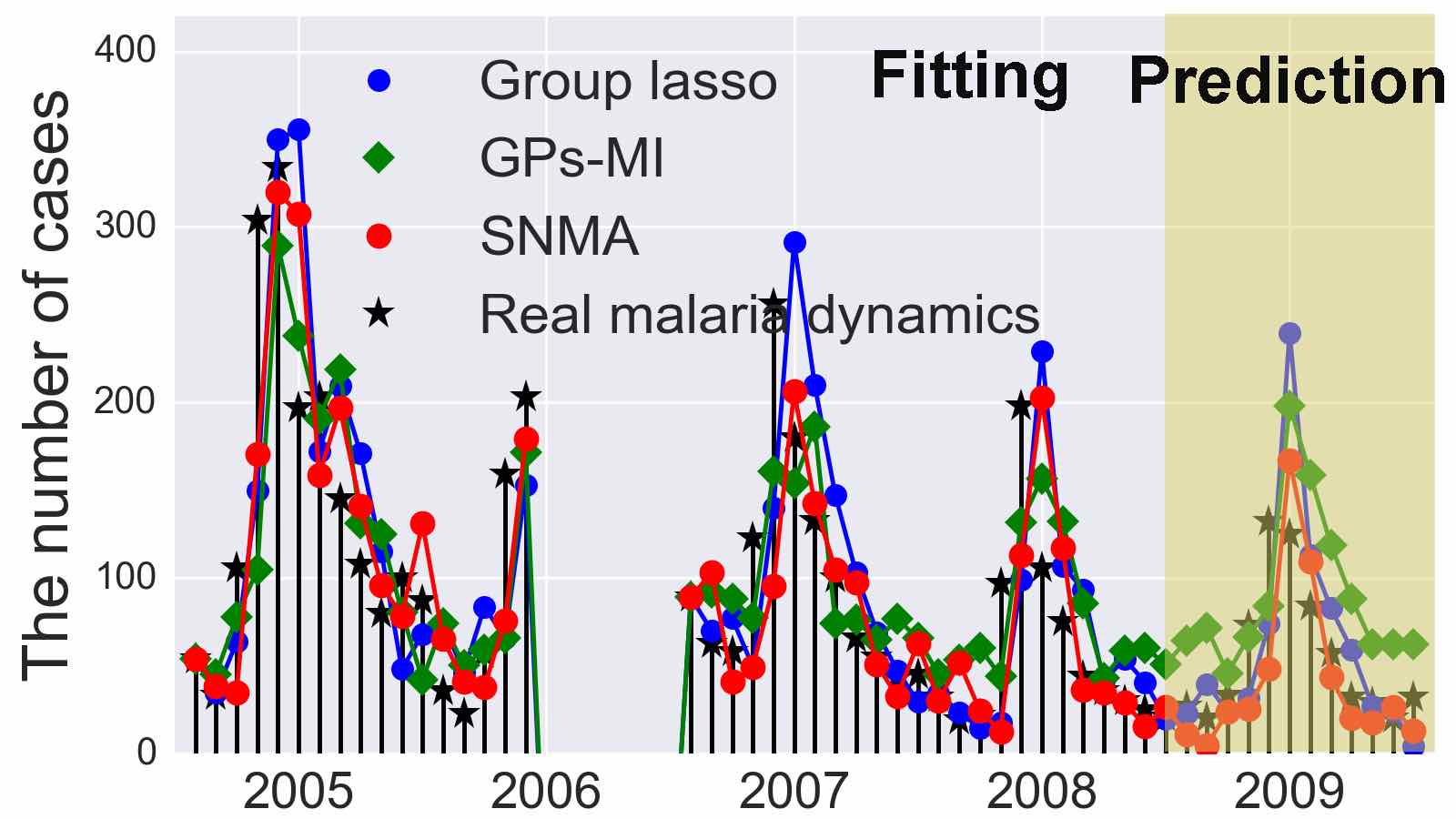} &
\includegraphics[width=0.31\textwidth]{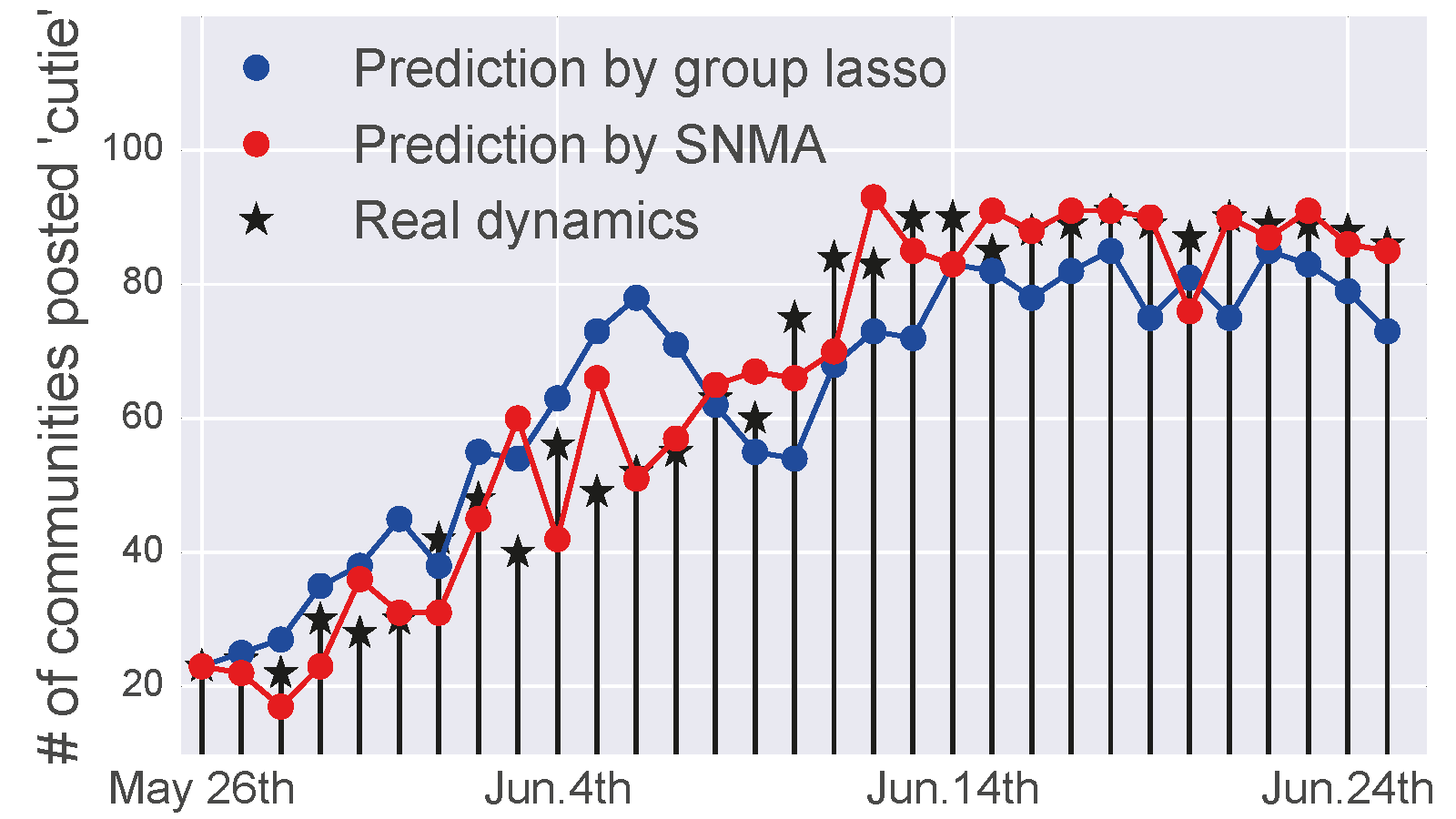}\\
(a) & (b) & (c)
\end{tabular}
\caption{Model fittings and sentinel predictions for epidemic dynamics. (a) 2009 Hong Kong H1N1 flu pandemic. (b) 2005-2009 Tengchong malaria. (c) The dynamics of hot word "cutie" in the Baidu Tieba. }
\label{fig:real_result}
\end{figure*}

Fig. \ref{fig:running_time} presents the average running time of one iteration of the three methods on a PC with a 3.4GHz CPU and 8GB memory.
Since GPs-MI employ forward greedy strategy (fast when only a few sentinels) but SNMA use backward selection method (fast when many sentinels), it's  fair to evaluate them by comparing the running time of one iteration of each algorithm.
The results show group lasso is the slowest, and GPs-MI and SNMA are almost at the same level.

\subsection{4.2 ~ Validations on real diffusion data}
In real cases, the failure rate cannot be evaluated because the ground truth sentinel network is unknown.

\subsubsection{4.2.1 ~ 2009 Hong Kong H1N1 flu pandemic}
The cases report data of 2009 Hong Kong H1N1 influenza epidemic, was provided by Centre for Health Protection (CHP), Department of Health, Government of the Hong Kong Special Administrative Region.
During this pandemic, the first imported case of human swine influenza (HSI) was confirmed on May 1, 2009.
As of Sep. 2010, there were over 36,000 confirmed cases of HSI, among which about 290 were severe cases and over 80 of them died \cite{CHPreports}.
We consider the epidemic dynamics for 105 days since the disease onset in Jun. 1st 2009 based on the confirmed cases of H1N1 infection reported by CHP (www.chp.gov.hk), which gives the spatial position and infection times of each case.

Hong Kong consists of 18 administrative districts, i.e., 18 components.
By merging the confirmed cases during 105 days, 3 days as a basic infectious period, we obtain the dynamics of $N$=$18$ and $T$=$35$.
We set the dynamics from Jun.1 to Aug.15 as training data and the one from Aug. 15 to Sep. 15 as test data.
According to a pre-given $k$, sentinel districts of Hong Kong can be identified via the methods (SNMA use the linear system setting).
Then, based on the sentinels, we predict the newly cases on test data as shown in Fig \ref{fig:sentinel_pre} (a). 
Obviously, SNMA outperforms the competitors.
Fig. \ref{fig:real_result} (a) shows a case of fittings and predictions for the dynamics ($8$ districts are selected as sentinels). The sentinel network of Hong Kong and the sentinels' spatial distribution can be found in Supporting Information.

\subsubsection{4.2.2 ~ 2005-2009 Tengchong malaria outbreak}
Tengchong City, Yunnan Province, China, has 18 towns, and 658,207 residents that are distributed in a wide area of 5,845 km$^2$ in 2011.
Because of the suitable climate for mosquito habitats, Tengchong has a quite serious malaria outbreak.
Five years' (2005-2009) monthly malaria cases data at the town level, were collected by Tengchong CDC and can be obtained from the annual reports of National Institute of Parasitic Disease, China CDC.
By eliminating the missing data from 2005 Jun. to Dec., we get malaria dynamics, which contains $N=18$ components and $T=53$ months.

We set the malaria dynamics from 2005 to 2008 as training data and the one during 2009 as test data.
Similar to the Hong Kong case, we identify the sentinel towns according to a pre-given $k$ and predict the dynamics on the test data, as shown in Fig \ref{fig:sentinel_pre} (b). 
Once again, SNMA achieves the best sentinel prediction in most cases.
Fig. \ref{fig:real_result} (b) shows a case of fittings and predictions for the malaria dynamics with $7$ sentinels. The sentinel network of Thengchong and the sentinels' distribution is in Supporting Information.

\subsubsection{4.2.3 ~ Hot words diffusion in Baidu Tieba}
Baidu Tieba (tieba.baidu.com), one of the largest online community platforms in China, is a collection of thousands of active topic-specific communities.
Tieba users can post any hot words (vocabularies that are widely used in Tieba during a short period) in any communities.
Hot words often present a diffusion phenomena in Baidu Tieba: a hot word first appears in only a few forums, and then is posted gradually in many other forums by the users who are active in multiple forums.
Thus, Tieba can be regarded as a logistical discrete dynamical system, where communities are components, hot words are contagions, and the infected state of a component is $0$ or $1$. 
GPs-MI cannot be applied to this case because it's base on Gaussian model and cannot directly work on discrete data.

We tracked the dynamics of 11 independent hot words cascading among the top-100 active communities in Baidu Tieba from Apr. 2014 to Oct. 2015 (18 months).
Only the dynamics during the words' bursting period is preserved, and the total bursting period of the words is 738 d, i.e., $N$ = $100$, $T$=$738$ by using 1 d as a time-unit. 
Here, we split the training and test data in term of the  hot words, i.e.,  we alternatively set the dynamics of one hot word as test data and the rest be the training data.
Then, we identify the sentinel communities in Baidu Tieba by the methods (GPs-MI cannot be applied in discrete data; SNMA use the logistic system setting).
Fig \ref{fig:sentinel_pre} (c) shows the results of sentinel prediction.
Fig. \ref{fig:real_result} (c) shows that the dynamics of the hot word ``cutie'' is successfully predicted based on the data from $66$ sentinel communities.

Summarily, SNMA outperform GPs-MI and group lasso in the most experiments.
Our method has two obvious advantages. 
(1) Our method is model-based  and can readily integrate prior knowledge, which makes it more effective and easier to train.
(2) Our method is more robust against noise and insufficient data owing to the Bayesian framework that can effectively handle the uncertainty from both data and model.

\section{5 ~ Conclusions}

In this work, we addressed the challenge of epidemic dynamics prediction in cases in which surveillance resources are very limited.
We proposed a novel importance measure, the $\bm{\gamma}$ value, by modeling a sentinel network with row sparse structure and presented an effective and flexible group sparse Bayesian learning algorithm for mining the sentinel network in two kinds of widely used dynamical systems.
With the discovered sentinel network, the overall epidemic dynamics can be predicted based on partial data only collected by the few sentinels.
Moreover, we significantly reduced the computational complexity of the algorithm and extended it to various nonlinear systems using basic function embedding technology.
We validated the proposed framework by a set of experiments on both synthetic and real-world datasets.



\section{Supporting Information}
\subsection{Theorem Proofs}
\subsubsection{Theorem 1}

\begin{proof}

\textbf{\emph{Linear continuous system.} }

The posterior distribution over $\bm{s}$ of the linear system is given by:
\begin{equation}  
p(\bm{s}|\bm{y}, \bm{\Phi}, \bm{\gamma}, \lambda)=\frac
{p(\bm{y}|\bm{\Phi}, \bm{s}, \lambda)  p(\bm{s}| \bm{\gamma}) }
{p(\bm{y}|\bm{\bm{\Phi},  \gamma}, \lambda)},
\label{si:cal_posterior_linear}
\end{equation} 
where $p(\bm{y}|\bm{\Phi}, \bm{s}, \lambda) \sim \mathcal{N}(\bm{\Phi} \bm{s},\lambda \mathbf{I})$ and $p(\bm{s}| \bm{\gamma}) \sim \mathcal{N}(\mathbf{0}, \bm{\Sigma}_0)$ (refer to Eq. \ref{equ:likelihood_con} and \ref{equ:prior_vector} in the main paper).
Thus, the numerator of Eq. \ref{si:cal_posterior_linear} is a product of two Gaussians, which is still a Gaussian.

The denominator is the marginal likelihood,
 $p(\bm{y}|\bm{\Phi}, \bm{\gamma}, \lambda) = \int p(\bm{y}|\bm{\Phi}, \bm{s}, \lambda)  p(\bm{s}| \bm{\gamma}) d\bm{s}$,
 which is a Gaussian convolution. 
The convolution can be analytically calculated and its result is also a Gaussian.
Then the posterior of $\bm{s}$ can be obtained, which is still a Gaussian, $p(\bm{s}|\bm{y}, \bm{\Phi}, \bm{\gamma}, \lambda) \sim \mathcal{N}(\bm{\mu}_{\bm{s}} , \bm{\Sigma}_{\bm{s}})$ with parameters
\begin{equation} \nonumber
\setlength{\abovedisplayskip}{3pt}
\setlength{\belowdisplayskip}{3pt}
\left\{
\begin{aligned}
 & \bm{\mu}_{\bm{s}}   = \lambda^{-1} \bm{\Sigma}_{\bm{s}} \bm{\Phi}^{\rm T} \bm{y}  \\
 & \bm{\Sigma}_{\bm{s}}^{-1}  = \bm{\Sigma}_0^{-1} + \lambda^{-1} \bm{\Phi}^{\rm T} \bm{\Phi}.
\end{aligned}
\right.
\end{equation}

\emph{\textbf{Logistic discrete system.}}

The posterior over $\bm{s}$ in the logistic system is given by:
\begin{equation} 
p(\bm{s}|\bm{y}, \bm{\Phi}, \bm{\gamma})=\frac
{p(\bm{y}|\bm{\Phi}, \bm{s})  p(\bm{s}| \bm{\gamma}) }
{p(\bm{y}|\bm{\Phi}, \bm{\gamma})},
\label{cal_posterior_logis}
\end{equation} 
where $p(\bm{y}|\bm{\Phi}, \bm{s})  p(\bm{s}| \bm{\gamma})$ is a Bernoulli distribution and $p(\bm{y}|\bm{\Phi}, \bm{\gamma})$ a Gaussian distribution (refer to Eq. \ref{likelihood_nonlinear}  and Eq. \ref{equ:prior_vector}  in the main paper).
We cannot directly calculate the Eq. \ref{cal_posterior_logis} because its denominator, the marginal likelihood $p(\bm{y}|\bm{\Phi}, \bm{\gamma})$ =  $\int p(\bm{y}|\bm{\Phi}, \bm{s})  p(\bm{s}| \bm{\gamma}) d\bm{s}$, cannot be analytically integrated.

We deduce the posterior in an alternative way. 
Firstly, by using the local variational method \cite{bishop2006pattern}, 
a lower bound function on the sigmoid having a Gaussian form can be introduced,
\begin{equation}  \nonumber
\sigma[\bm{\Phi}_n \bm{s}]  \geq   \sigma[\xi_n] e^{ \frac{1}{2}(\bm{\Phi}_n \bm{s} - \xi_n) - \bm{\pi} (\xi_n) ((\bm{\Phi}_n \bm{s})^{2} - \xi_{n}^{2})},
\label{xi_low_bound}
\end{equation}
where $\bm{\pi} (\xi_n)$ = $- \frac{1}{2\xi_n}(\sigma [\xi_n] - \frac{1}{2})$ and $\xi_n$ is an introduced variational parameter. 

By substituting the lower bound into likelihood function (refer to Eq. \ref{likelihood_nonlinear} in the main paper), the likelihood of the logistic system $p(\bm{y}|\bm{\Phi}, \bm{s} )=\prod \nolimits_{n=1}^{TN} \sigma[\bm{\Phi}_{n} \bm{s}]^{\bm{y}_n} (1- \sigma[\bm{\Phi}_{n}\bm{s}] )^{1- \bm{y}_n}$,
we get a lower bound on the likelihood,
\begin{equation} \nonumber
\begin{split}
p(\bm{y}|\bm{\Phi}, \bm{s} ) & \geq h(\bm{s}, \bm{\xi})\\
 =  \prod \nolimits_{n=1}^{TN} & \sigma[\xi_n] e^{\bm{\Phi}_n \bm{s} \bm{y}_n - \frac{1}{2}(\bm{\Phi}_n \bm{s} + \xi_n) - \bm{\pi} (\xi_n) ((\bm{\Phi}_n \bm{s})^{2} - \xi_{n}^{2}) }.
\end{split}
\label{low_bound_theorem}
\end{equation}

Then we have a lower bound on the joint distribution of $\bm{s}$ and $\bm{y}$:
\begin{equation}
\begin{aligned} 
 & p(\bm{y}, \bm{s}|\bm{\Phi}, \bm{\gamma}) =  p(\bm{y}|\bm{\Phi}, \bm{s})  p(\bm{s}| \bm{\gamma})  \geq  h(\bm{s}, \bm{\xi}) p(\bm{s}| \bm{\gamma}) \\
 & =  e^{ \frac{1}{2}\bm{s}^{\rm T} \Sigma_0 ^{-1} \bm{s} + \sum_{n=1} ^{TN} (\bm{\Phi}_n \bm{s} (\bm{y}_n - \frac{1}{2}) - \bm{\pi}(\xi_n) (\bm{\Phi}_n \bm{s})^2 ) + const }.
 \label{joint_dis}
\end{aligned}
\end{equation}
Note that, the exponent term in Eq. \ref{joint_dis} is a quadratic function of $\bm{s}$. 
By identifying the linear and quadratic terms of $\bm{s}$ and then normalizing the function over $\bm{s}$, we obtain a Gaussian approximation to the original posterior over $\bm{s}$, $p(\bm{s}|\bm{y}, \bm{\Phi}, \bm{\gamma}) \sim \mathcal{N}(\bm{\mu}_{\bm{s}} , \bm{\Sigma}_{\bm{s}})$ with parameters
\begin{equation} \nonumber
\left\{
\begin{aligned}
&  \bm{\mu}_{\bm{s}} = 2^{-1} \bm{\Sigma}_{\bm{s}} \bm{\Phi}^{\rm T}(2\bm{y} - \mathbf{1})\\
& \bm{\Sigma}_{\bm{s}}^{-1} = \bm{\Sigma}_0^{-1} + \bm{\Phi}^{\rm T} \bm{\pi} (\bm{\xi}) \bm{\Phi},
\end{aligned}
\right.
\end{equation}
where vector $\bm{\xi}$= $[\xi_1, \dots, \xi_{TN}]^{\rm T}$, $\bm{\pi} (\bm{\xi}) $ $\in$ $\mathbb{R}^{TN \times TN}$ is a diagonal matrix. Specially, $\bm{\pi} (\bm{\xi})_{n,n}$ = $\bm{\pi} (\xi_n)$ =  $- \frac{1}{2\xi_n}(\sigma [\xi_n] - \frac{1}{2})$.

\end{proof}

\subsubsection{Theorem 2}

\begin{proof}
Without loss of generality, let $\mathbf{A}$ $\in$ $\mathbb{R}^{nk \times mk}$ and $\mathbf{B}$ $\in$ $\mathbb{R}^{mk \times lk}$ denote two diagonal-block matrix. 
The sub-matrix of $\mathbf{A}$ is $\mathbf{A}_{ij}$ $\in$ $\mathbb{R}^{k \times k}$, who is a diagonal matrix with all diagonal entries having the same value entry $a_{ij}$. 
Analogously, matrix $\mathbf{B}$,  sub-matrix $\mathbf{B}_{ij}$, and entry $b_{ij}$ have the same relationship.

Based on the property of block matrix, the product $\mathbf{C}$ $\in$ $\mathbb{R}^{nk \times lk}$ is a block matrix and its sub-matrix $\mathbf{C}_{ij}$ =$\sum_{r=1}^{m} \mathbf{A}_{ir} \mathbf{B}_{rj}$, for $i$=$1:n, j$=$1:l$.
And $\mathbf{C}_{ij}$  is a diagonal matrix with all diagonal entries having the same value $c_{ij}$=$\sum_{r=1}^{m} a_{ir} b_{rj}$.
Thus, $\mathbf{C}$ is a diagonal-block matrix according to the Definition 1 (in the main paper), and $\bar{\mathbf{C}}$ is the block projective matrix of $\mathbf{C}$ according to the Definition 2 (in the main paper).

For each entry of $\bar{\mathbf{C}}$  $\in$ $\mathbb{R}^{n \times l}$, 
$\bar{\mathbf{C}}_{ij}$ = $c_{ij}$ =$\sum_{r=1}^{m} a_{ir} b_{rj}$= $\sum_{r=1}^{m} \bar{\mathbf{A}}_{ir} \bar{\mathbf{B}}_{rj}$.
Therefore, $\bar{\mathbf{C}}$  = $\bar{\mathbf{A}}$ $\bar{\mathbf{B}}$.
\end{proof}

\vspace{3mm}
\subsubsection{Theorem 3}

\begin{proof}
Without loss of generality, let square matrix $\mathbf{A}$ $\in$ $\mathbb{R}^{nk \times nk}$ denote a diagonal-block matrix.
The adjugate matrix of $\mathbf{A}$,  $\rm{adj}(\mathbf{A})$, is also a diagonal-block matrix because that if an entry is zero in the matrix $\mathbf{A}$ the corresponding entry who has the same location in adjugate matrix $\rm{adj}(\mathbf{A})$ must be also zero.

The inverse matrix of $\mathbf{A}$ can be calculated by
\begin{equation} \nonumber
\mathbf{A}^{-1} = \frac{1}{\rm{ det}( \mathbf{A}) } \rm{adj}(\mathbf{A}),
\end{equation}
where the determinant $\rm{ det}( \mathbf{A})$ is a scalar. 
Thus the inverse matrix $\mathbf{A}^{-1}$ has the same structure with $ \rm{adj}(\mathbf{A})$, i.e., the inverse matrix $\mathbf{A}^{-1}$ is also a diagonal-block matrix.

Let $\bar{\mathbf{A}}$ $\in$ $\mathbb{R}^{n \times n}$ be the block projective matrix of square matrix $\mathbf{A}$. 
Based on eigendecomposition rules, we have $\mathbf{A}$ = $\mathbf{Q}_1 \bm{\Lambda}_1 \mathbf{Q}^{-1}_1$ and $\bar{\mathbf{A}}$ = $\mathbf{Q}_2 \bm{\Lambda}_2 \mathbf{Q}_2^{-1}$, where $\bm{\Lambda}_1$ and $\bm{\Lambda}_2$ are two diagonal matrices whose diagonal elements are the corresponding eigenvalues, i.e., $\bm{\Lambda}_1$=diag$(\bm{\lambda}_1)$ and  $\bm{\Lambda}_2$=diag$(\bm{\lambda}_2)$, where vector $\bm{\lambda}_1$ and $\bm{\lambda}_2$ are the eigenvalues of $\mathbf{A}$ and $\bar{\mathbf{A}}$, respectively.

Let $\bm{\lambda}_{2}(i)$ denote the $i$-th eigenvalue in $\bm{\lambda}_2$. 
We have $\bar{\mathbf{A}} \bm{v}_2(i)$ = $\bm{\lambda}_2(i) \bm{v}_2(i)$, where $\bm{v}_2(i)$ is the corresponding eigenvector.
According to the Theorem 2, we have $\mathbf{A} \bm{v}_1(i)$ = $\bm{\lambda}_2(i) \bm{v}_1(i)$, where $\bm{v}_1(i)$ $\in$ $\mathbb{R}^{nk \times k}$ is a diagonal-block matrix and its block projective matrix is $\bm{v}_2(i)$.
That is to say, $\bm{\lambda}_2(i)$ is a $k$-tuply repeated eigenvalue to $\mathbf{A}$, i.e., the entries of $\bm{\lambda}_2$ repeat $k$ times in $\bm{\lambda}_1$.
Therefore, $\bm{\Lambda}_2$ is the block projective matrix of $\bar{\bm{\Lambda}}_1$, i.e., $\bar{\bm{\Lambda}}_1$ = $\bm{\Lambda}_2$. 
As  $\bm{\Lambda}_1$ and $\bm{\Lambda}_2$ are both diagonal matrices, their inverse matrices satisfy $\bar{(\bm{\Lambda}_1^{-1})}$ = $\bm{\Lambda}_2^{-1}$.
Analogously, we can also have $\bar{\bm{Q}}_1$ = $\bm{Q}_2$.

 The inverse matrix of $\mathbf{A}$ and $\bar{\mathbf{A}}$ are $\mathbf{A}^{-1}$ = $\mathbf{Q}_1 \bm{\Lambda}^{-1}_1 \mathbf{Q}_1^{-1}$ and   $(\bar{\mathbf{A}})^{-1}$ = $\mathbf{Q}_2 \bm{\Lambda}^{-1}_2 \mathbf{Q}_2^{-1}$, respectively.
Based on the conclusions of $\bar{\bm{Q}}_1$ = $\bm{Q}_2$ and $\bar{(\bm{\Lambda}_1^{-1})}$ = $\bm{\Lambda}_2^{-1}$, we have $\bar{(\mathbf{A}^{-1})}$ = $(\bar{\mathbf{A}})^{-1}$ according to the Theorem 2.
\end{proof}

\subsection{Derivation of Hyper-parameters Learning Rules}

We propose two group sparse Bayesian learning methods to mine sentinel network for linear continuous systems and logistic discrete systems, respectively.
We estimate the hyper-parameters set $\Theta$ by employing the EM (Expectation-Maximization) method to maximize the marginal likelihood $p(\bm{y}|\bm{\Phi} ,\Theta)$, by treating $\bm{s}$ as hidden variables.
Specifically, $\Theta=\{ \lambda, \bm{\gamma} \}$ for the linear system, and $\Theta=\{\bm{\gamma}, \bm{\xi} \}$ for the logistic system.

\subsection{Linear continuous system}
Based on the posterior (refer to the Theorem 1 in the main paper), the the conditional expectation of complete log-likelihood, i.e., the Q-function of the linear system, is given by,
 \begin{equation} \nonumber
  \begin{split}
Q(\bm{\gamma}, \lambda) & = E_{\bm{s}| \bm{y}, \bm{\Phi},\bm{\gamma}^{\rm (old)} , \lambda^{\rm (old)}}
[\log (p(\bm{y}|\bm{\Phi}, \bm{s}, \lambda)  p(\bm{s}| \bm{\gamma}) )] \\
& =E_{\bm{s}| \bm{y}, \bm{\Phi}, \bm{\gamma}^{\rm (old)} , \lambda^{\rm (old)}}[\log p(\bm{y}| \bm{\Phi}, \bm{s}, \lambda) + \log p(\bm{s}| \bm{\gamma}) ]
\end{split}
\label{Q_fucction_con}
 \end{equation}
where $E_{\bm{s}| \bm{y}, \bm{\Phi}, \bm{\gamma}^{\rm (old)} , \lambda^{\rm (old) }}[\cdot]$ denotes the conditional expectation with respect to the posterior, and $\cdot ^{\rm (old) }$ denotes the parameter estimated in the previous iteration.
Notice that the log-terms of $\bm{\gamma}$ and $\lambda$ are separated in $Q(\bm{\gamma} , \lambda)$. 
Thus we can divide the Q-function into two simplifying sub-Q-functions. 

The sub-Q-funciton for $\bm{\gamma}$ is:
\begin{equation} \nonumber
\begin{split}
Q(\bm{\gamma}) =& E_{\bm{s}| \bm{y}, \bm{\Phi},\bm{\gamma}^{\rm (old)} , \lambda^{\rm (old)}}
[\log (p(\bm{s}| \bm{\gamma}))]\\
\propto & -\log |\bm{\Sigma}_0| - {\rm Tr} [\bm{\Sigma}_0^{-1}(\bm{\Sigma}_{\bm{s}}+\bm{\mu}_{\bm{s}} \bm{\mu}_{\bm{s}}^{\rm T})].
\label{q_fun_lambda}
\end{split}
\end{equation}

The derivative of $Q(\bm{\gamma})$ with respect to $\gamma_{i}$ is given by,
\begin{equation}  \nonumber
\frac{\partial Q(\bm{\gamma})}{\partial \bm{\gamma}_i} = -\frac{N}{2 \bm{\gamma}_i} + \frac{1}{2 \bm{\gamma}_i^2} {\rm Tr}[\bm{\Sigma}_{\bm{s}}^i+\bm{\mu}_{\bm{s}}^i  (\bm{\mu}_{\bm{s}}^i)^{\rm T}],
\end{equation}
where  $\bm{\mu}_{\bm{s}}^i$ $\in$ $\mathbb{R}^{N \times 1}$  and $\bm{\Sigma}_{\bm{s}}^i$ $\in$ $\mathbb{R}^{N \times N}$ are the $i$-th block in $\bm{\mu}_{\bm{s}}$ and $\bm{\Sigma}_{\bm{s}}$ respectively.

By setting the derivative be zero, the update rule for $\gamma_i$, i.e., the $\bm{\gamma}$-value of component $i$, can be obtained (the Eq. \ref{update_gamma} in the main paper),
\begin{equation} \nonumber
\gamma_i \leftarrow ((\bm{\mu}_{\bm{s}}^i)^{\rm T}  \bm{\mu}_{\bm{s}}^i + {\rm Tr}[ \bm{\Sigma}_{\bm{s}}^i ])N^{-1}, ~  i=1,\dots,N.\label{si:update_gamma}
\end{equation}

To estimate $\lambda$, the sub-Q-funciton is given by
\begin{equation}  \nonumber
\begin{split}
Q(\lambda) & \propto  - TN \log \lambda - \frac{1}{\lambda} E_{\bm{s}| \bm{y},\bm{\Phi},  \bm{\gamma}^{(\rm old)} , \lambda^{(\rm old)}}[ \parallel \bm{y}-\bm{\Phi} \bm{s}  \parallel_2^2 ] \\
 = &- TN \log \lambda - \frac{1}{\lambda} ( \parallel \bm{y}-\bm{\Phi} \bm{\mu}_{\bm{s}}  \parallel_2^2 \\
& + E_{\bm{s}| \bm{y},\bm{\Phi},  \bm{\gamma}^{(\rm old)} , \lambda^{(\rm old)}}[ \parallel \bm{\Phi} (\bm{s}  - \bm{\mu}_{\bm{s}}   )  \parallel_2^2 ])  \\
 =& - TN \log \lambda - \frac{1}{\lambda} ( \parallel \bm{y}-\bm{\Phi} \bm{\mu}_{\bm{s}}  \parallel_2^2 + {\rm Tr} [\bm{\Sigma_{\bm{s}}} \bm{\Phi}^{\rm T} \bm{\Phi} ] ) 
\label{q_fun_lambda}
\end{split}
\end{equation} 

The derivative of the $Q(\lambda)$ with respect to $\lambda$ is given by
\begin{equation}  \nonumber
\frac{\partial Q(\lambda)}{\partial \lambda} = -  \frac{TN}{\lambda}  + \frac{1}{\lambda^2} ( \parallel \bm{y}-\bm{\Phi} \bm{\mu}_{\bm{s}}  \parallel_2^2 + {\rm Tr} [\bm{\Sigma_{\bm{s}}} \bm{\Phi}^{\rm T} \bm{\Phi} ] )
\end{equation} 

By setting the derivative to zero, the update rule for $\lambda$ can be obtain (the Eq. \ref{update_lambda} in the main paper),
\begin{equation}  \nonumber
\lambda \leftarrow (TN)^{-1} (\parallel \bm{y}-\mathbf{\Phi} \mathbf{\mu}_{\bm{s}}  \parallel_2^2 + {\rm Tr} [\bm{\Sigma_{\bm{s}}} \mathbf{\Phi}^{\rm T} \mathbf{\Phi} ])
\label{si:update_lambda}
\end{equation}

\subsection{Logistic discrete system}
As aforementioned the marginal likelihood in the posterior of $\bm{s}$ cannot be analytically integrated in the logistic system, we utilize local variational method to estimate the hyper-parameters in the EM framework.
Firstly, we introduce a lower bound function of sigmoid function,
\begin{equation} \nonumber
\sigma[\bm{\Phi}_n \bm{s}]  \geq   \sigma[\xi_n] e^{ \frac{1}{2}(\bm{\Phi}_n \bm{s} - \xi_n) - \bm{\pi} (\xi_n) ((\bm{\Phi}_n \bm{s})^{2} - \xi_{n}^{2})},
\label{xi_low_bound}
\end{equation}
where $\bm{\pi} (\xi_n)$ = $- \frac{1}{2\xi_n}(\sigma [\xi_n] - \frac{1}{2})$ and $\xi_n$ is an introduced variational parameter. 

By substituting the lower bound function into the likelihood of the logistic system (refer to Eq. \ref{likelihood_nonlinear} in the main paper), a lower bound on the likelihood of the logistic system is given by:
\begin{equation} \nonumber
\begin{split}
p(\bm{y}|\bm{\Phi}, \bm{s} ) & \geq h(\bm{s}, \bm{\xi})\\
 =  \prod \nolimits_{n=1}^{TN} & \sigma[\xi_n] e^{\bm{\Phi}_n \bm{s} \bm{y}_n - \frac{1}{2}(\bm{\Phi}_n \bm{s} + \xi_n) - \pi (\xi_n) ((\bm{\Phi}_n \bm{s})^{2} - \xi_{n}^{2}) },
\end{split}
\label{low_bound}
\end{equation}
where, vector $\bm{\xi}$ $\in$ $\mathbb{R}^{TN}$ is the introduced variational parameters and its entry is $\xi_n$. Then the Q-function can be written as:
 \begin{equation} \nonumber
Q(\bm{\xi}, \bm{\gamma} ) = E_{\bm{s}| \bm{y}, \mathbf{\Phi},\bm{\gamma}^{\rm (old)} , \bm{\xi}^{\rm (old)}}
[\log (h(\bm{s}, \bm{\xi})  p(\bm{s}| \bm{\gamma}) )].
\label{Q_fucction_dis}
 \end{equation}
This Q-function also can be separated w.r.t $\bm{\gamma}$ and $\bm{\xi}$, and the sub-Q-funciton for $\bm{\xi}$ is:
\begin{equation} \nonumber
Q(\bm{\xi})  \propto \sum  _{n=1} ^{TN}  \frac{\log \sigma[\xi_n] - \xi_n}{2} - \pi (\xi_n) (\bm{\Phi}_n (\Sigma_{\bm{s}} +\bm{\mu}_{\bm{s}} \bm{\mu}_{\bm{s}}^{\rm T} ) \bm{\Phi}_n ^{\rm T} - \xi_n^2 ).
\label{q_fun_xi}
\end{equation}
By letting the derivative of $Q (\bm{\xi})$ over $\xi_n$ to be zero,  we can get the learning rule for $\xi_n$ (the Eq. \ref{update_xi} in the main paper):
\begin{equation}  \nonumber
\xi_n \leftarrow \sqrt{ \bm{\Phi}_n (\Sigma_{\bm{s}} +\bm{\mu}_{\bm{s}} \bm{\mu}_{\bm{s}}^{\rm T} ) \bm{\Phi}_n ^{\rm T}}, ~ n = 1,\dots,TN.
\label{si:update_xi}
\end{equation}

In the logistic system, the update rule for $\bm{\gamma}$ is the same as the linear system (the Eq. \ref{update_gamma} in the main paper), as they have the same sub-Q-funciton of $\bm{\gamma}$.

\subsection{Supporting Figures}

\begin{figure}[h]
\centering{}
\includegraphics[width=7cm]{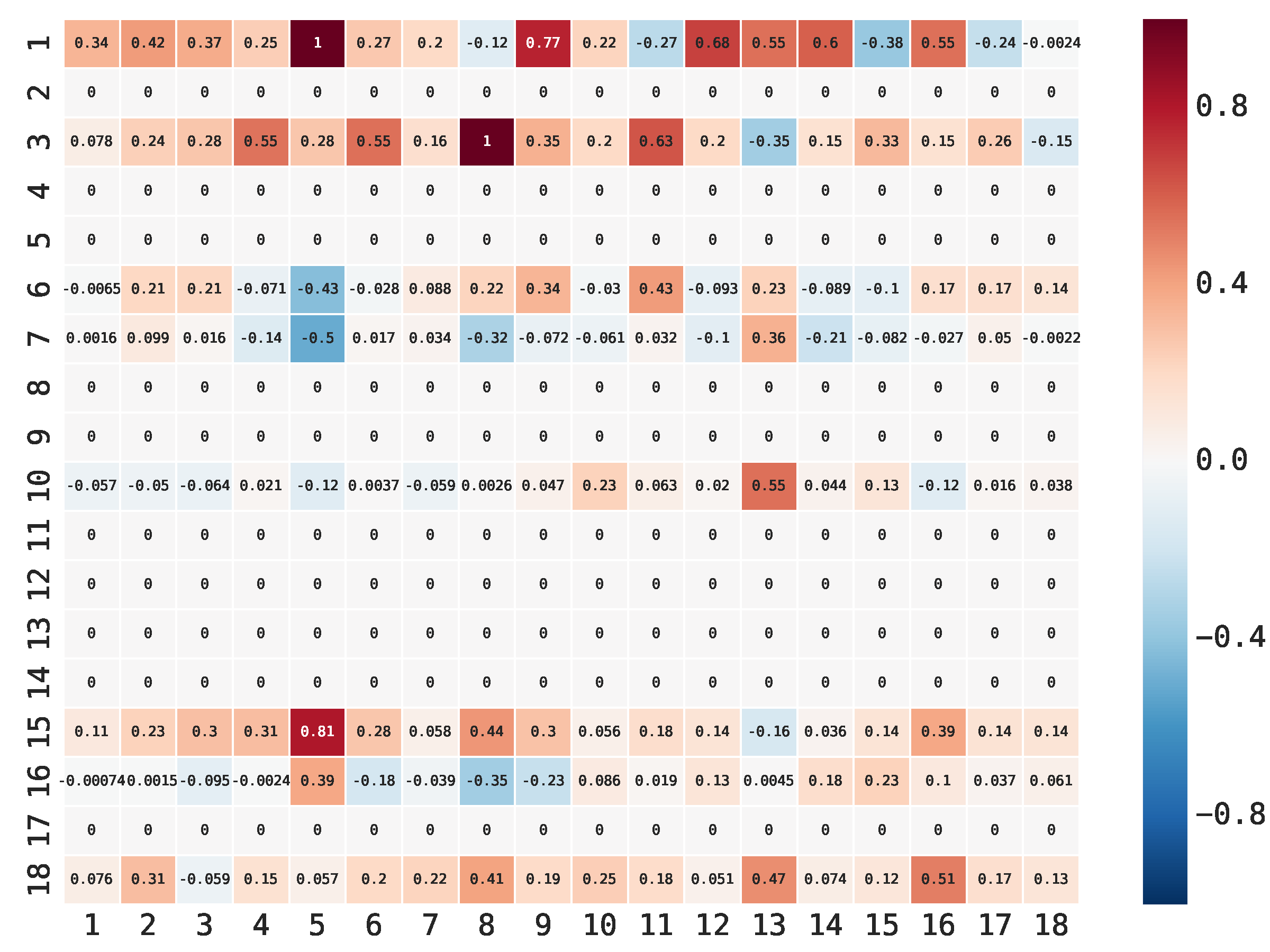}
\caption{The sentinel network of Hong Kong inferred from 2009 Hong Kong H1N1 flu dynamics, where 7 dense rows indicate 7 sentinels.}
\label{fig:running_time}
\end{figure}

\begin{figure}[h]
\centering{}
\includegraphics[width=7cm]{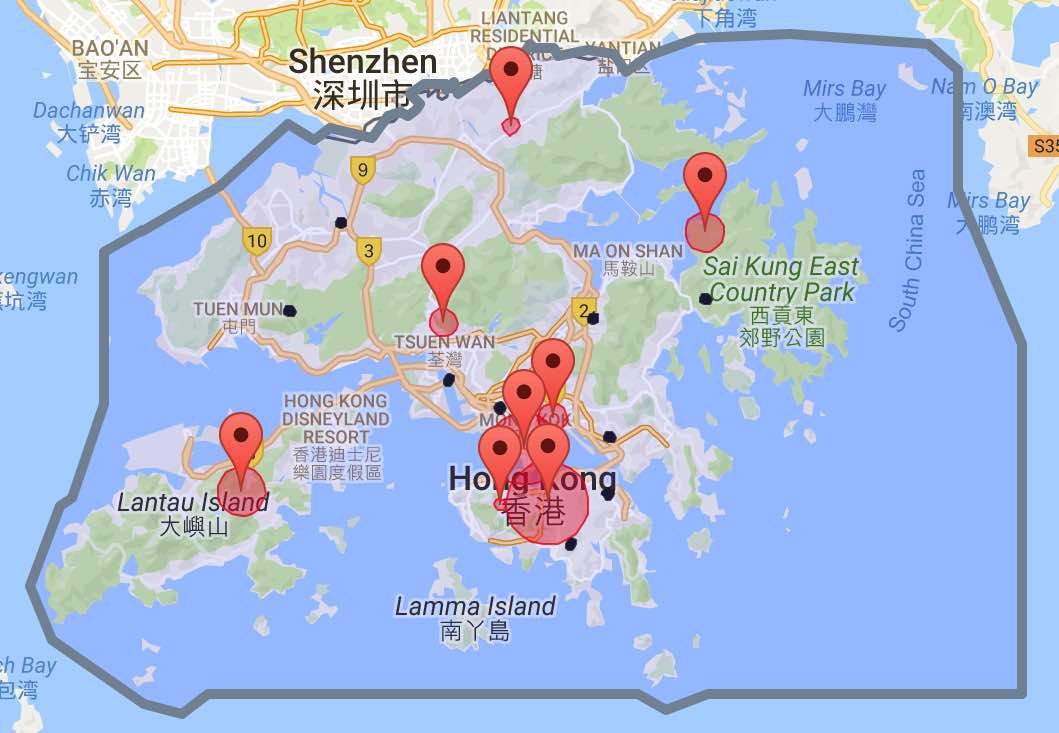}
\caption{The spatial distribution of the 8 sentinel districts ($44\%$ districts) in Hong Kong. The red bubble markers denote the sentinel locations, and the radius of red circle depicts its importance for dynamics prediction. The black points are unmonitored locations.}
\end{figure}

\begin{figure}[h]
\centering{}
\includegraphics[width=7cm]{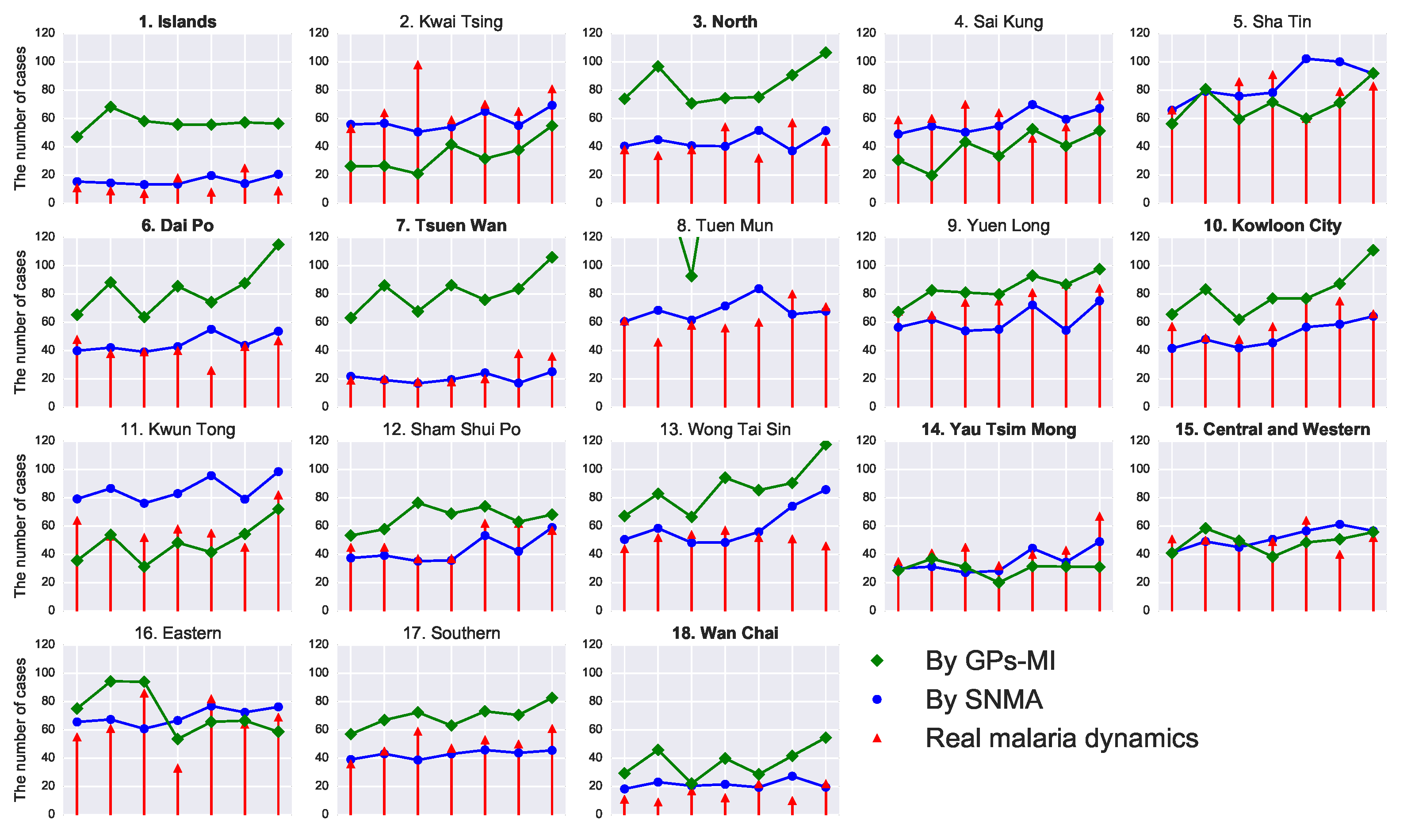}
\caption{Prediction comparison between GPs-MI and SNMA on district level. Red stems denote the real dynamics of H1N1 flu in the 18 districts of Hong Kong from Aug. 15 to Sep. 15 (the testing data of Hong Kong experiments in main paper). The blue line and green line denote the predictions by SNMA and GPs-MI, respectively}
\end{figure}

\begin{figure}[h]
\centering{}
\includegraphics[width=7cm]{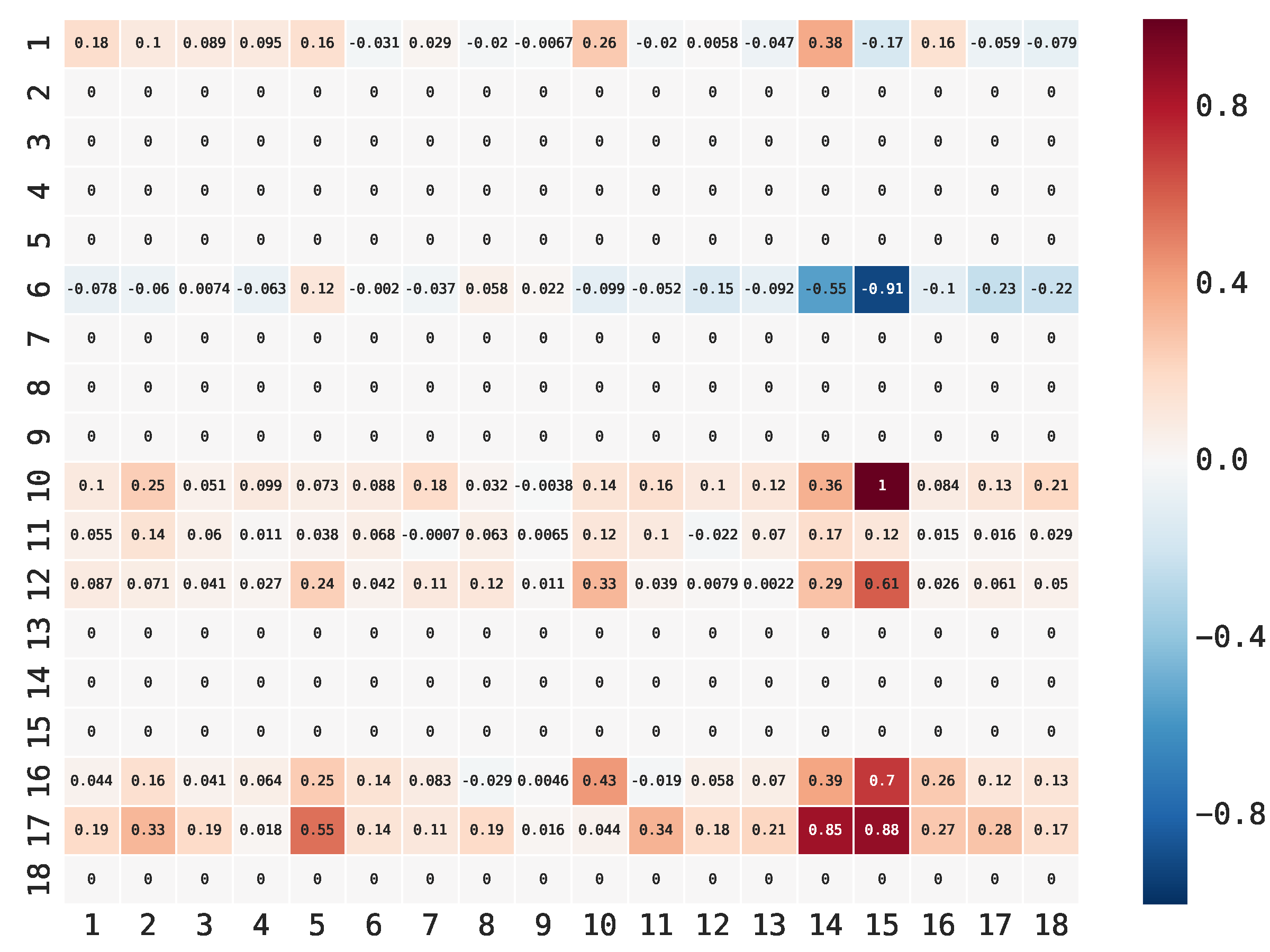}
\caption{The sentinel network  of Tengchong inferred from 2005-2009 Tengchong malaria dynamics, where 8 dense rows indicate 8 sentinels.}
\end{figure}

\begin{figure}[h]
\centering{}
\includegraphics[width=4.5cm]{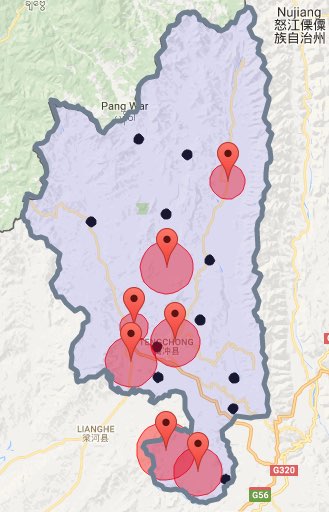}
\caption{The spatial distribution of the 8 sentinel towns ($38\%$ towns) in Tengchong city. The red bubble markers denote the sentinel locations, and the radius of red circle depicts its importance for dynamics prediction. The black points are unmonitored locations.}
\end{figure}

\begin{figure}[h]
\centering{}
\includegraphics[width=7cm]{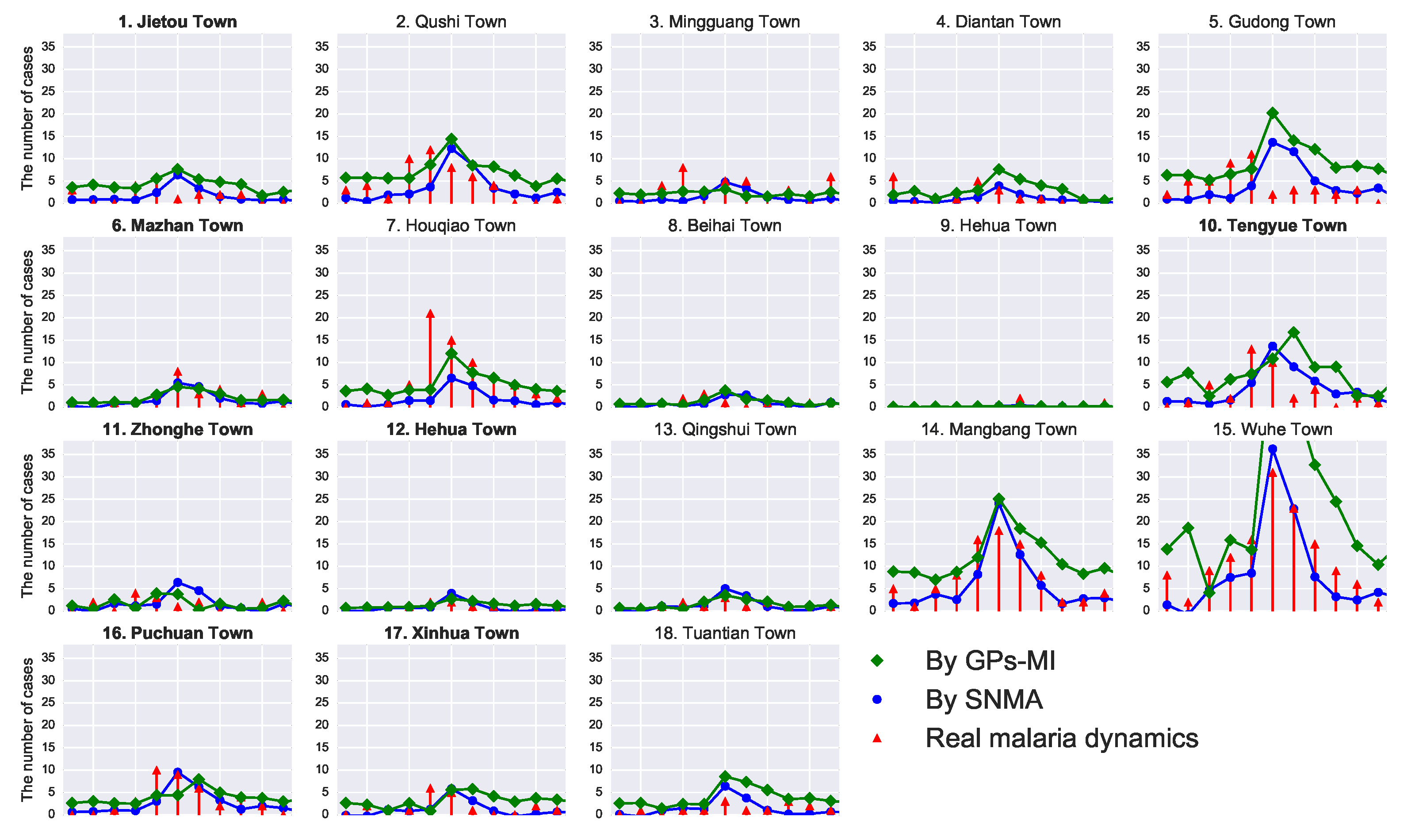}
\caption{Prediction comparison between GPs-MI and SNMA on town level. Red stems denote the real dynamics of malaria in the 18 towns of Tengchong during 2009 (the testing data of Tengchong experiments in main paper). The blue line and green line denote the predictions by SNMA and GPs-MI, respectively}
\end{figure}

\begin{figure}[h]
\centering{}
\includegraphics[width=7cm]{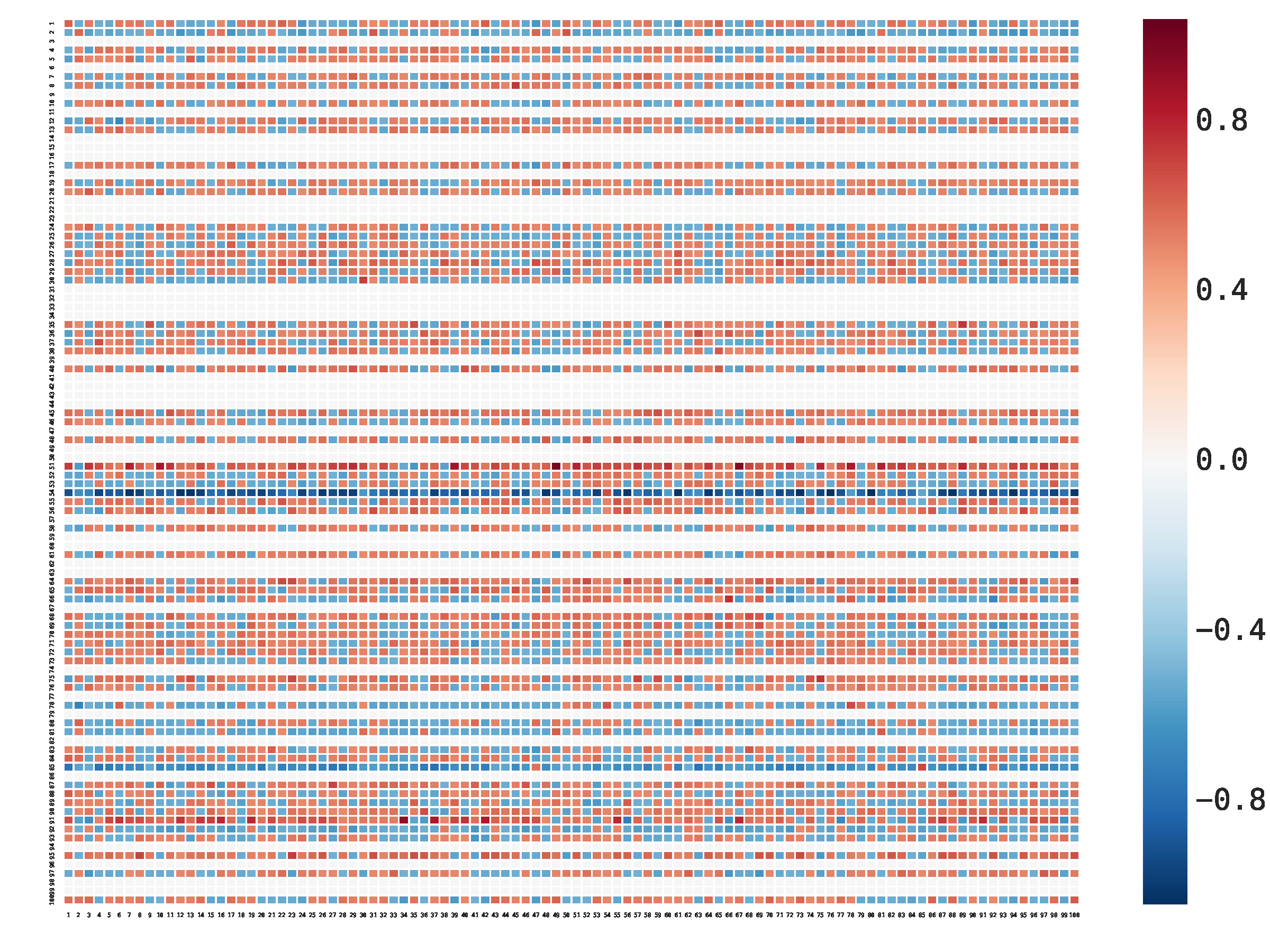}
\caption{The sentinel network  of Baidu Tieba inferred from 10 independent hot words' cascading among the top-100 active communities in Baidu Tieba from Apr. 2014 to Oct. 2015 (18 months). The 66 dense rows indicate 66 sentinel communities.}
\end{figure}

\clearpage
\bibliographystyle{aaai}
\bibliography{reference} 

\end{document}